\documentclass[10pt,twocolumn,letterpaper]{article}

\usepackage[pagenumbers]{cvpr} 

\usepackage{graphicx}
\usepackage{listings}
\usepackage{booktabs}
\usepackage[T1]{fontenc}
\usepackage{amssymb}
\usepackage{amsmath}
\usepackage{algorithm}
\usepackage{algpseudocode}
\usepackage{comment}
\usepackage{tikz}
\usepackage{pgfplots}
\pgfplotsset{compat=1.18}
\usepackage{enumitem}

\definecolor{cvprblue}{rgb}{0.21,0.49,0.74}

\def\paperID{9302} 
\def\confName{CVPR}
\def\confYear{2025}

\usepackage[pagebackref,breaklinks,colorlinks,allcolors=cvprblue]{hyperref}
\usepackage{mdframed}
\usepackage[capitalize]{cleveref}

\title{Certified Human Trajectory Prediction}

\author{Mohammadhossein Bahari$^{*,1}$ \quad Saeed Saadatnejad$^{*,1}$ \quad Amirhossein Askari Farsangi$^{1}$
\\
Seyed-Mohsen Moosavi-Dezfooli$^{\dagger,2}$ \quad Alexandre Alahi$^{1}$
\\\\
$^{1}$EPFL \quad $^{2}$Apple \\
{\tt\small {\{mohammadhossein.bahari, saeed.saadatnejad\}}@epfl.ch} \\}

\begin{document}
\maketitle
\let\thefootnote\relax\footnotetext{\leftline{$^* $ Equal contribution. }}
\let\thefootnote\relax\footnotetext{\leftline{$^\dagger $ Work done while at Imperial College London.}}


\begin{abstract}
Predicting human trajectories is essential for the safe operation of autonomous vehicles, yet current data-driven models often lack robustness in case of noisy inputs such as adversarial examples or imperfect observations. Although some trajectory prediction methods have been developed to provide empirical robustness, these methods are heuristic and do not offer guaranteed robustness.
In this work, we propose a certification approach tailored for trajectory prediction that provides guaranteed robustness. To this end, we address the unique challenges associated with trajectory prediction, such as unbounded outputs and multi-modality. To mitigate the inherent performance drop through certification, we propose a diffusion-based trajectory denoiser and integrate it into our method. Moreover, we introduce new certified performance metrics to reliably measure the trajectory prediction performance. Through comprehensive experiments, we demonstrate the accuracy and robustness of the certified predictors and highlight their advantages over the non-certified ones.
The code is available online: \href{https://s-attack.github.io/}{https://s-attack.github.io/}  
\end{abstract}



\section{Introduction}

Predicting the behavior of humans is a crucial task for the safe operation of autonomous vehicles and robots. The task, known as human trajectory prediction, aims to predict the future positions of humans given their past positions. It has received significant attention in recent years, with data-driven methods demonstrating remarkable performance~\cite{girgis2022autobot,xu2023eqmotion,saadatnejad2024socialtransmotion}.
Such progress prompts a critical question: \textit{Are these methods reliable enough for real-world applications with noisy inputs?} Notably, it has been shown that these methods are susceptible to adversarial attacks, raising significant concerns regarding their robustness and security~\cite{saadatnejad2021sattack,cao2022advdo,tan2023targeted}.
Moreover, recent findings indicate that in real-world autonomous driving pipelines, inputs of the prediction models are imperfect, resulting in performance drops~\cite{xu2024towards}.
This noise stems from upstream modules in the perception pipeline, such as detection and tracking.
Therefore, it is crucial to study the robustness properties of trajectory predictors.

\begin{figure}[!t]
    \centering
    \includegraphics[width=\columnwidth]{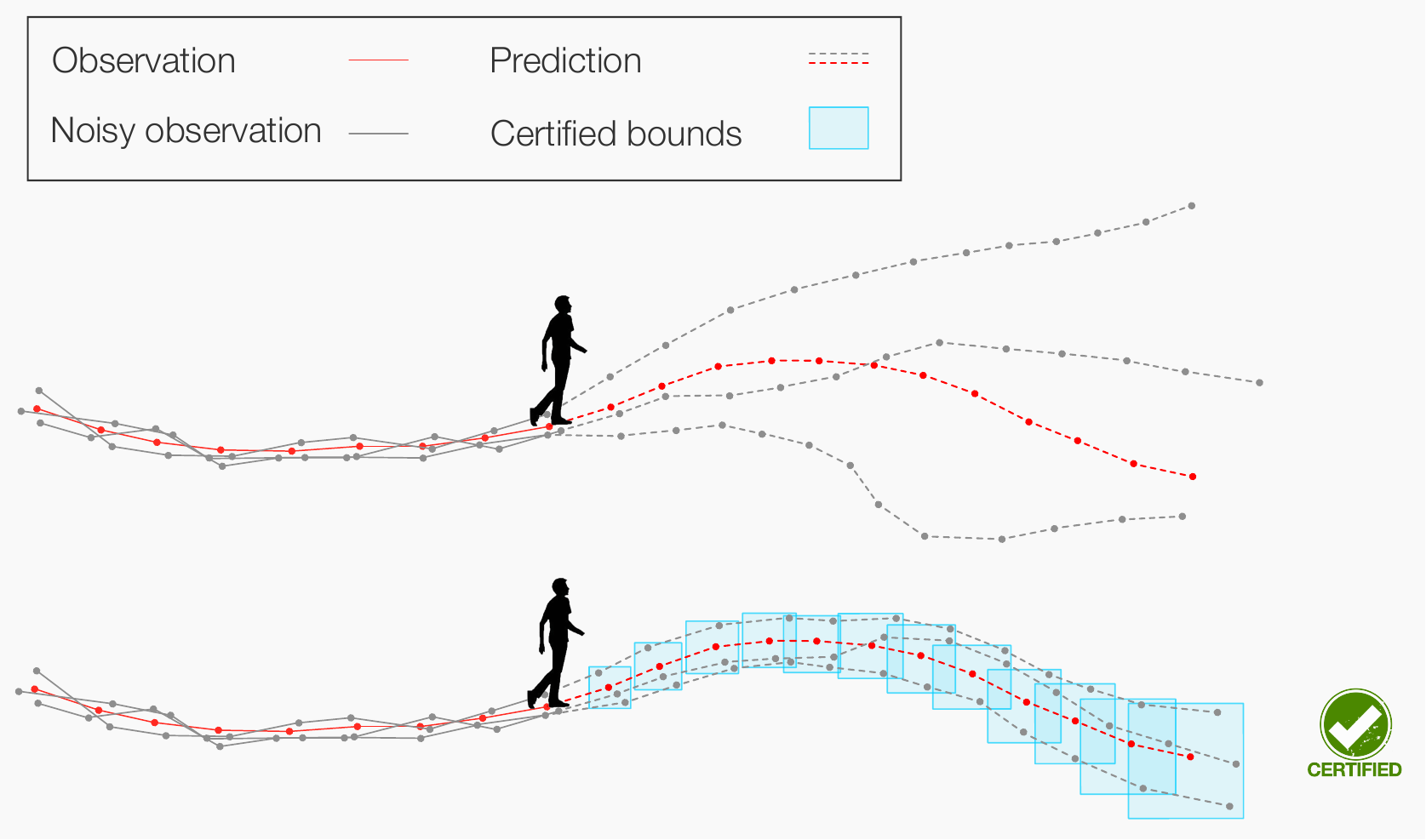}
    \caption{Illustration of the influence of noisy inputs on trajectory prediction models. 
    The red trajectories depict clean observations and the corresponding predictions, while the gray trajectories show noisy observations along with their resulting predictions. The top part showcases the outputs of a standard trajectory prediction model, revealing unbounded predictions with noisy inputs. In contrast, the bottom part demonstrates the outputs of our trajectory predictor with guaranteed robustness. The model provides certified bounds (blue boxes) on the predicted outputs, ensuring that outputs remain within guaranteed regions despite input noise. 
    }
    \label{fig:pull} 
\end{figure}

Previous works proposed heuristic approaches to improve the empirical robustness of the trajectory prediction models~\cite{zhang2022adversarial,cao2023robust,jiao2023semi}.
However, it has been shown that such heuristic approaches are ultimately ineffective against sufficiently powerful adversaries~\cite{carlini2017adversarial,uesato2018adversarial,athalye2018obfuscated}. 
Therefore, it is essential to study certification techniques that provide \textit{guaranteed} robustness  \textit{i.e.,} to guarantee that given any confined input noise, the models' outputs fall within certified bounds. Given the black-box nature of the prediction models, the bounds on the outputs also deliver reliability to the system which is crucial for autonomous vehicles. 
We illustrate in \Cref{fig:pull} that even slight perturbations in observation inputs considerably changes model's predictions. Nevertheless, predictions from a model with guaranteed robustness invariably stay within its certified bounds.

In this work, we propose a certification method for trajectory prediction based on ``Randomized Smoothing''~\cite{cohen2019certified_kolter}.
In principle, randomized smoothing transforms a base model into a smoothed model by adding random perturbations to the input, and then aggregating the outputs. 
Up to our knowledge, we are the first to study this in trajectory prediction, encountering various challenges: 
(1)~How can we transform the randomized smoothing technique widely applied in image classification to the multi-output regression task of trajectory prediction?
Moreover, it is a time-series regression task that does not inherently have a maximum output range, which is essential for randomized smoothing. How can we define a range for the outputs? 
(2)~While randomized smoothing is known to hurt performance in classification~\cite{cohen2019certified_kolter}, to what extent does it hurt the accuracy of trajectory predictors? Is there a way to maintain the accuracy?
(3)~Finally, trajectory predictors are often multi-modal, generating multiple output modes. How can these multi-modal outputs be accommodated in the certification process? 
In order to address the aforementioned challenges, 
(1)~we adapt two randomized smoothing approaches based on mean~\cite{cohen2019certified_kolter} and median~\cite{chiang2020detection_as_regression} aggregation functions to trajectory prediction and compare their performances. 
We also propose an adaptive clamping strategy to pose maximum output ranges. 
(2)~To mitigate the performance degradation resulting from randomized smoothing, we propose a denoiser as a pre-processor suppressing the noise before feeding to the smoothed model. Given the success of diffusion models, we design an unconditional diffusion-based denoiser tailored for trajectory data. 
(3)~Finally, we address the multi-modality challenge by proposing a new certification definition that can accommodate multi-modal outputs. 

We conduct experiments employing state-of-the-art trajectory prediction models trained on Trajnet++ benchmark~\cite{kothari2021human}.
We develop smoothed trajectory prediction models with guaranteed robustness and demonstrate both their accuracy and the certified bounds of their predictions. The results highlight the advantages of the smoothed models over non-certified models in noisy input settings. They also indicate that the most accurate models are not necessarily the most robust. 
In addition, we show that common performance metrics for the trajectory prediction task are unreliable as they cannot account for the potential input noises. 
To address this, we introduce new certified metrics, equipped with the certified bounds.

In summary, our contributions are as follows:

\begin{itemize}[noitemsep, topsep=0pt, leftmargin=*]
\item We are the first to introduce certification to the trajectory prediction task, providing guaranteed robustness for models against adversarial attacks and imperfect inputs.
\item We develop a randomized smoothed trajectory predictor 
tailored to the unique challenges of the task and propose an unconditional diffusion denoiser to enhance the performance.
\item We introduce new certified performance metrics and through comprehensive experiments, demonstrate the accuracy and robustness of the smoothed models and highlight their advantages over non-certified models. 
\end{itemize}

\section{Related Works}
\label{sec:rel}

\textbf{Human trajectory prediction.}
In recent years, as autonomous driving systems and social robots have become more popular, the challenge of predicting human trajectories has caught much attention. The majority of the research revolves around enhancing accuracy by learning the interaction dynamics between humans more effectively. To this end, Social-LSTM~\cite{alahi2016sociallstm} is the pioneering work employing neural networks. Subsequent studies propose different architectural solutions based on Convolutional Neural Networks~\cite{nikhil2018convolutional,zamboni2022pedestrian}, Graph Neural Networks~\cite{cao2021spectral,mohamed2020social}, and Transformers~\cite{giuliari2021transformer,franco2023under, saadatnejad2024socialtransmotion,lee2024mart,saadatnejad2024toward}.
Additional approaches have incorporated the domain knowledge~\cite{kothari2021interpretable,liu2021socialnce}, developed equivarient feature learning~\cite{girgis2022autobot,xu2023eqmotion} and explored various strategies for pooling social features~\cite{kothari2021human,bartoli2018context,ivanovic2019trajectron}.

\noindent\textbf{Robustness for trajectory prediction.}
The vulnerability of trajectory predictors to adversarial attacks has been shown in several previous works~\cite{saadatnejad2021sattack,cao2022advdo,tan2023targeted}. To address this vulnerability, others proposed robustness defenses based on various heuristic approaches~\cite{bahari2022sattack,zhang2022adversarial,cao2023robust,jiao2023semi}. However, none of these approaches are guaranteed robustness methods. 
Recently, Trajpac~\cite{zhang2023trajpac} proposed a verification approach for the robustness of trajectory predictors. They employ a probably approximately correct (PAC) strategy by approximating the prediction model locally with a linear model and use it as a proxy to determine the robustness of the model.
However, their method has some limitations: 
(1) Their method is not agnostic to the input noise distribution due to the dependency of learned linear model on the noise distribution fed during learning. 
(2)  Their method is inefficient in the number of required samples, with experiments often necessitating over $30,000$ random samples. 
(3) Their method is probabilistic, and does not provide a guaranteed robustness. 
In this work, we employ a randomized smoothing approach that provides a certified bound, requires significantly fewer samples, and generalizes to any noise distribution encountered during deployment. 

\noindent\textbf{Randomized smoothing certification.}
Certification is to guarantee that a model’s outputs are within a bound around its initial output once the model's inputs are within a neighborhood of its initial input and is mainly used as a defense against adversarial attacks. 
Various certification and verification methods
have been proposed based on Satisfiability Modulo theories~\cite{ehlers2017formal,huang2017safety}, mixed integer linear programming~\cite{bunel2018unified,fischetti2018deep}, solving optimization problems~\cite{wong2018provable,dvijotham2018dual} and layer by layer outer approximation of activations~\cite{singh2018fast}. 
However, these methods are computationally expensive and cannot scale to common neural networks. Alternatively, randomized smoothing has been proposed as an efficient and model-agnostic approach and has achieved great success in the classification task~\cite{cao2017mitigating,liu2018towards,carlini2022certified_diffusion}. More importantly, it imposes the least assumptions on the noise distribution, providing robustness against any confined noise.  
In randomized smoothing, the smoothed prediction for a given input is calculated by sampling some points around that input and aggregating their corresponding outputs. ~\citet{cohen2019certified_kolter} proved certified bounds for the smoothed predictors with a mean aggregator, particularly for the classification task. Moreover, it was shown that integrating a denoiser into a smoothed predictor can greatly enhance both accuracy and certified bounds~\cite{salman2020denoised}. Recently, randomized smoothing certification has been adapted for the detection task~\cite{chiang2020detection_as_regression}. It introduces a median smoothing aggregator which is more appropriate for regression tasks.
To the best of our knowledge, our work is the first  randomized smoothing certification for the trajectory prediction problem, studying both mean and median smoothing. 

Randomized smoothing is distinct from other methods that guarantee models' output such as 
conformal prediction~\cite{shafer2008tutorial}
as conformal prediction provides the guarantee of ground truth coverage rather than the guarantee of the output region. Moreover, unlike randomized smoothing, conformal prediction is dependent on the input noise distribution (calibration set).
Randomized smoothing is also different from uncertainty quantification approaches~\cite{hullermeier2021aleatoric} and they serve distinct yet complementary purposes. 
Uncertainty estimation quantifies the model's uncertainty (aleatoric or epistemic) for a given input, while randomized smoothing transforms the original model's outputs into a new output with bounds through smoothing.

\noindent\textbf{Denoising diffusion.}
Denoising diffusion probabilistic models~\cite{ho2020ddpm} have achieved great success in image generation~\cite{rombach2022high,gal2022image,zhang2023adding,khachatryan2023text2video}, human pose prediction~\cite{saadatnejad2023diffusion}, GPS trajectory generation~\cite{zhu2023difftraj} and even trajectory prediction~\cite{gu2022stochastic,mao2023leapfrog,bae2024singulartrajectory,wang2024optimizing}. However, the application of these models as denoisers remains largely underexplored with only a few studies investigating their use as denoisers in other domains such as image restoration~\cite{yang2023real, zhu2023denoising}. In training a diffusion model, Gaussian noise is progressively added to the input during the forward process. The model is then trained to reverse this process, recovering the input over several steps. 
This makes the diffusion model particularly suitable for denoising tasks on any noisy signal. 
 We are the first to propose an unconditional diffusion-based denoiser for trajectory data and integrate it into our randomized smoothed predictor.

\section{Method}

\label{sec:method}
In this section, we first explain the certification framework backgrounds and then describe our certification for the trajectory prediction task.

\begin{figure*}[!t]
    \centering
    \includegraphics[width=0.95\textwidth]{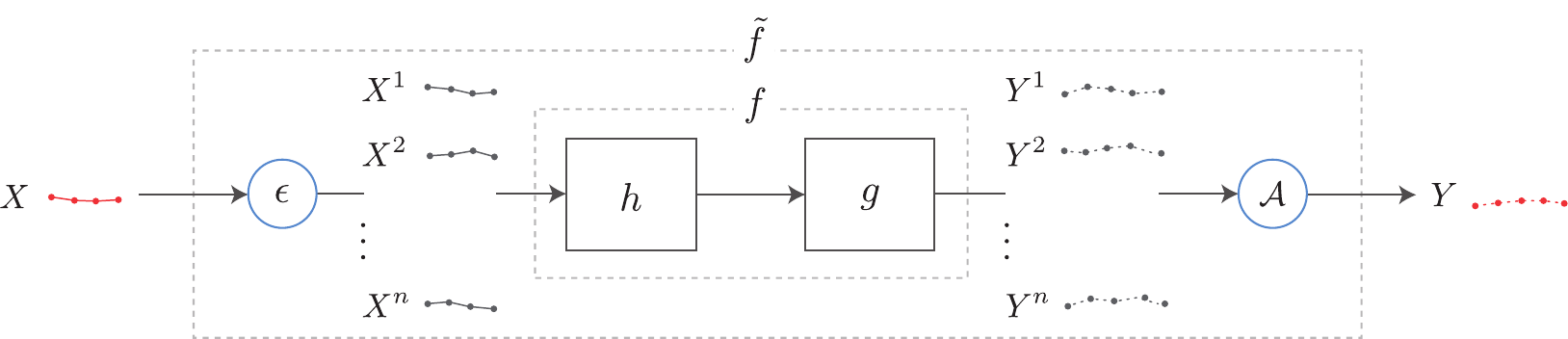}
    \caption{An outline of the proposed smoothed trajectory predictor: $n$ different randomized input observations $X^i$ are created by adding perturbation $\epsilon$ to the input $X$. The denoiser $h$ processes these samples $X^i$, which are then fed into the trajectory predictor $g$ to make the outputs $Y^i$. Applying an aggregation function $\mathcal{A}$ (median or mean) on $Y^i$, the final smoothed prediction $Y$ is derived.}
    \label{fig:network}
\end{figure*}

\subsection{Certification}
\label{sec:background}

Randomized smoothing~\cite{cohen2019certified_kolter} is a technique initially introduced for certifying the robustness of models against $\ell_2$-norm adversarial attacks in image classification. 
Given a prediction function $f$
, randomized smoothing aims to bound the output of a smoothed function $\tilde{f}=\mathcal{A}(f)$ where $\mathcal{A}$ is an aggregation/smoothing operator. This bound is valid for a radius in the neighborhood of the input, named certification radius $R$.
 
We consider two choices for the smoothing operator: mean~\cite{cohen2019certified_kolter} and median~\cite{chiang2020detection_as_regression} smoothing.

\paragraph{Mean smoothing.} Given a function $f: \mathbb{R}^d \rightarrow [l, u]$, input $X\in \mathbb{R}^d$, and input certification radius $R$, mean smoothing computes the expected value of the predictor over a perturbed input, that is  $ \tilde{f}(X) = \mathbb{E}_\epsilon[f(X+\epsilon)],$
where $\epsilon \sim N\left(0, \sigma^2 I\right)$. It has been shown that given
$\lVert r \rVert_2 < R$, the output of $\tilde{f}$ can be bounded as:
\begin{equation}
\begin{split}
    &l + (u - l) \cdot \Phi \left(\frac{\eta(X) - R}{\sigma} \right) \leq \tilde{f}(X+r),\\
    &\tilde{f}(X+r)\leq l + (u - l) \cdot \Phi \left(\frac{\eta(X) + R}{\sigma} \right),
\label{eq:bounds1}
\end{split}
\end{equation}
where $\eta(X) = \sigma\cdot\Phi^{-1}\left(\frac{\tilde{f}(X)-l}{u-l}\right)$ and $\Phi$ is the cumulative distribution function of the standard Gaussian. We refer to the lower and upper certified bounds as LB and UB, respectively.

Trajectory predictors are commonly multi-output $f: \mathbb{R}^d \rightarrow [l_1, u_1] \times [l_2, u_2] \times \cdots \times [l_m, u_m]$, where $f(X) = (f_1(X), \ldots, f_m(X))$. In this case, the certification bounds are applicable individually to each coordinate. 
We will explain the estimation process for $l_i$'s and $u_i$'s in \Cref{sec:certrap}.

Note that this smoothing method is applicable to functions with initial lower and upper bounds. However, for functions that inherently lack those, an alternative option is to use median smoothing.

\paragraph{Median smoothing.} Given a continuous function $f: \mathbb{R}^d \rightarrow \mathbb{R}^m$, and an input certification radius $R$, median smoothing aims to find a bound for the median of predictions, as given by $\tilde{f}(X) = q_{0.5}(X),$
where
$q_{p}(X) = \sup\{y \in \mathbb{R} \mid \mathbb{P}[f(X+\epsilon) \leq y] \leq p\}$
is the quantile function with $q_{0.5}$ indicating the median and $\epsilon \sim  N(0, \sigma^2 I)$.
Then, the certified bounds for $\|r\|_2\leq R$ are as follows:
\begin{align}
    q_{\Phi \left(-\frac{R}{\sigma} \right)}(X)
    \leq
    \tilde{f}(X+r)
    \leq 
    q_{\Phi \left(\frac{R}{\sigma} \right)}(X).
    \label{eq:bounds2}
\end{align}
Similar to the mean smoothing, we refer to the lower and upper certified bounds as LB and UB, respectively.
\begin{mdframed}
In simple words, with certification, we ensure that if an input to the smoothed predictor is perturbed within a radius $R$, the output remains within a certified range.
\end{mdframed}
Note that, while $R$ is a parameter determined by the application's requirements, $\sigma$ serves as a hyperparameter. As we will see later, $\sigma$ can be adjusted to balance between performance and the tightness of bounds.
For instance, when $\sigma=0$, the output aligns closely with the original predictor, 
and it yields trivial bounds $l\leq \tilde{f} \leq u$.
As $\sigma$ increases, the influence of the perturbation becomes more pronounced, causing the certified bounds to tighten, albeit with more smoothed/less accurate predictions. 
We will analyze the effect of $\sigma$ in \Cref{sec:exp}.

\subsection{Certified Trajectory Prediction}
\label{sec:certrap}

Human trajectory prediction tackles a regression task with sequences as inputs and outputs.
The position of an agent at any timestep $t$ is represented by its xy-coordinates $(x_t, y_t)$.
Given an observation sequence for \(T_{\text{obs}}\) timesteps as \(X=(x_{-T_{\text{obs}}+1}, y_{-T_{\text{obs}}+1}, \dots, x_0, y_0)\), the model predicts the next  \(T_{\text{pred}}\) positions \(Y=g(X)=(x_1, y_1, \dots, x_{T_{\text{pred}}}, y_{T_{\text{pred}}})\), aiming to be close to the ground-truth \(\hat{Y}\).
Notably, the trajectory predictor \(g\) can be construed as a function mapping \(\mathbb{R}^{2 T_{\text{obs}}} \rightarrow \mathbb{R}^{2 T_{\text{pred}}}\), making it suitable for certification purposes with \(d=2 T_{\text{obs}}\) and \(m=2 T_{\text{pred}}\).

\Cref{fig:network} provides an overview of our approach. 
Initially, we acquire $n$ Monte-Carlo samples from $\epsilon \sim N(0, \sigma^2 I)$, adding them to input $X$ to get $X^1, \cdots, X^n$. 
They are then processed by our denoiser $h$.
The certified bounds for $\tilde{f}(X)=\mathcal{A}(g(h(X+\epsilon)))$ are then computed according to \Cref{eq:bounds1} and \Cref{eq:bounds2} \footnote{Note that while $\epsilon$ has a Gaussian distribution, bounds are valid for any noise distribution within radius $R$.}. Note that $\mathcal{A}$ represents the aggregation function (either median or mean) applied to $Y^1, \cdots, Y^n$ to yield the final smoothed prediction $Y$. 
In the followings, we explain the details of the method, and defer the full algorithm to the supplementary.

\textbf{Diffusion denoiser.} The denoised smoothing technique combines a classifier with a denoiser, by first passing perturbed inputs through the denoiser to pre-process them before being fed into the model~\cite{salman2020denoised}. Extending this technique to trajectory prediction, we combine the predictor $g$ with $h$, making $f(X)=g(h(X))$. 
The denoiser suppresses the noise before feeding the data to the predictor, resulting in tighter certified bounds for the composed model $f$.
In an optimal scenario, where the denoiser exhibits high efficacy ($h(X+\epsilon) \approx X$), we obtain pseudo-clean data for $g$, leading to prediction performance closely resembling that of original data. 
When the denoiser is absent, we put $h=\mathrm{id}$ and the certification is hold for $f(X)=g(X)$.
We propose a simple autoencoder for $h$ designed to unconditionally denoise the input. This model is trained independently from the predictor through multiple steps of a denoising diffusion process, enabling it to learn the distribution of trajectory data. At inference time, the diffusion process is repeated for the required number of steps in order to denoise the input trajectory. We explain more our diffusion model in the supplementary.

\textbf{Adaptive clamping.} As mentioned in~\Cref{sec:background}, to establish certified bounds in case of mean aggregation for the multi-output human trajectory predictor with \(m=2 T_{\text{pred}}\) outputs, one needs to compute $l_j$'s and $u_j$'s.
However, the output of trajectory predictors inherently lacks bounds due to the unrestricted nature of the predicted positions. To address this for our certification equations, we propose adaptive clamping.
The process involves computing the predictions given all $X$ in the training dataset. By determining the maximum and minimum values from these computations, we establish $l_j=\min_{X}f_j(X)$ and $u_j=\max_{X}f_j(X)$ for each coordinate $j$.
However, these bounds are not guaranteed. In other words, with new samples, the predictor may predict outside these estimated bounds. Therefore, we cannot derive the certified bounds using the previous equations. 
To address this, all coordinates of the predicted trajectories, $f_j(X)$'s, are clamped with $\min(u_j, \max(l_j, \, . \, ))$ operator to ensure conformity within the specified range.
Note that one advantage of median smoothing is that the initial bounds are not required.

\textbf{Certified metrics.}
The smoothed predictor generates a predicted trajectory with a certified bound around each predicted timestep.
In order to assess them, we introduce the following metrics:

\begin{itemize}[noitemsep, topsep=0pt, leftmargin=*]
\item{\textbf{Average / Final Bound half-Diameter (ABD/FBD)}:}
    ABD measures the distance of the farthest points within the bound from the predicted trajectory, averaged over all timesteps, and FBD measures this distance at the final timestep as:
    \begin{equation}
    \begin{split}
    \text{FBD} = \frac{1}{2} \Big[ 
    &\left(\text{UB}_{2T_{\text{pred}}-1} - \text{LB}_{2T_{\text{pred}}-1}\right)^2 \\
    + &\left(\text{UB}_{2T_{\text{pred}}} - \text{LB}_{2T_{\text{pred}}}\right)^2 
    \Big]^{0.5}
    \end{split}
    \label{eq:fbd}
    \end{equation}

    \item \textbf{Certified-ADE / Certified-FDE:}
    The common Average/Final Displacement Error (ADE/FDE) metrics are typically reported under the assumption of perfect inputs. However, in practical scenarios, various types of input noise can occur, which can significantly alter the performance of predictors. 
    To address this gap, we propose these metrics that measure the highest ADE/FDE happening given noisy inputs. Specifically, they measure the distance of the farthest point inside the bounds to the ground-truth trajectory. 
    \item \textbf{Certified Collision Rate (Certified-Col):}
    Collision rate has been previously introduced as a metric that quantifies the percentage of collisions between the predicted trajectory of an agent and the ground-truth trajectories of neighboring agents in the scene~\cite{kothari2021human}. We introduce this metric as the percentage of examples in which at least one neighboring agent lies within the calculated certified bounds of the predicted trajectory.
\end{itemize}

\textbf{Multi-modality.} 
Unlike the classification task, the trajectory prediction is a multi-modal task\textit{ i.e.,} given an input trajectory, multiple plausible trajectories can be predicted as output. 
Nonetheless, it poses a unique challenge for certification in multi-modal predictors generating $k$ modes on which mode to consider. To address this, we reformulate it into multi-output mapping $\mathbb{R}^{2 T_{\text{obs}}} \rightarrow \mathbb{R}^{k \times 2 T_{\text{pred}}}$, and leverage the fact that each mode captures a specific behavior.
Consequently,  we certify all $k$ modes and choose the best mode based on the minimum of the Certified-FDE among all modes. 
The corresponding metrics are then computed for the selected mode. 

\section{Experiments}
\label{sec:exp}

\textbf{Datasets:}
ETH~\cite{pellegrini2010eth}, UCY~\cite{lerner2007ucy}, and WildTrack~\cite{chavdarova2018wildtrack} are well-established datasets containing annotations of human positions in crowded environments.
We utilize the Trajnet++~\cite{kothari2021human} benchmark, which provides a fixed data split and unified pre-processing for these datasets.
We used the common input and output lengths of \(T_{\text{obs}}=9\) and  \(T_{\text{pred}}=12\).

\textbf{Baselines:}  Up to our knowledge, our work is the first to introduce certification for trajectory prediction. Therefore, we conduct a comparative analysis between the mean and median smoothed predictors, representing the two certification techniques introduced. In order to obtain the smoothed predictors, we transform multiple existing trajectory predictors into smoothed models.   
We employ multiple state-of-the-art learning-based trajectory predictors, namely Directional-Pooling (D-Pool)~\cite{kothari2021human}, AutoBot~\cite{girgis2022autobot}, and EqMotion~\cite{xu2023eqmotion}. Moreover, we include Social-Force~\cite{helbing1998social-forces} as a rule-based trajectory predictor. 

\textbf{Metrics:} We report the performances in terms of the Average / Final Displacement Error (ADE / FDE) between the model predictions and its ground-truth values, along with FBD, Certified-FDE, and Certified-Col, introduced in \Cref{sec:method}. The reported values are in meters and percentages. For the sake of space, we leave the results on ABD and Certified-ADE for the supplementary.

\textbf{Implementation details:}
Throughout the experiments, the number of Monte-Carlo samples $n$ is set to $100$, $R$ to $0.1$, and $\sigma$ range to $0.08-0.4$. 
Note that since $\sigma$ serves as a hyperparameter, this range has been experimentally selected to ensure the models perform effectively. 

\subsection{Results}
\label{sec:results}

\begin{figure*}[!t]
    \centering
    \includegraphics[width=\columnwidth]{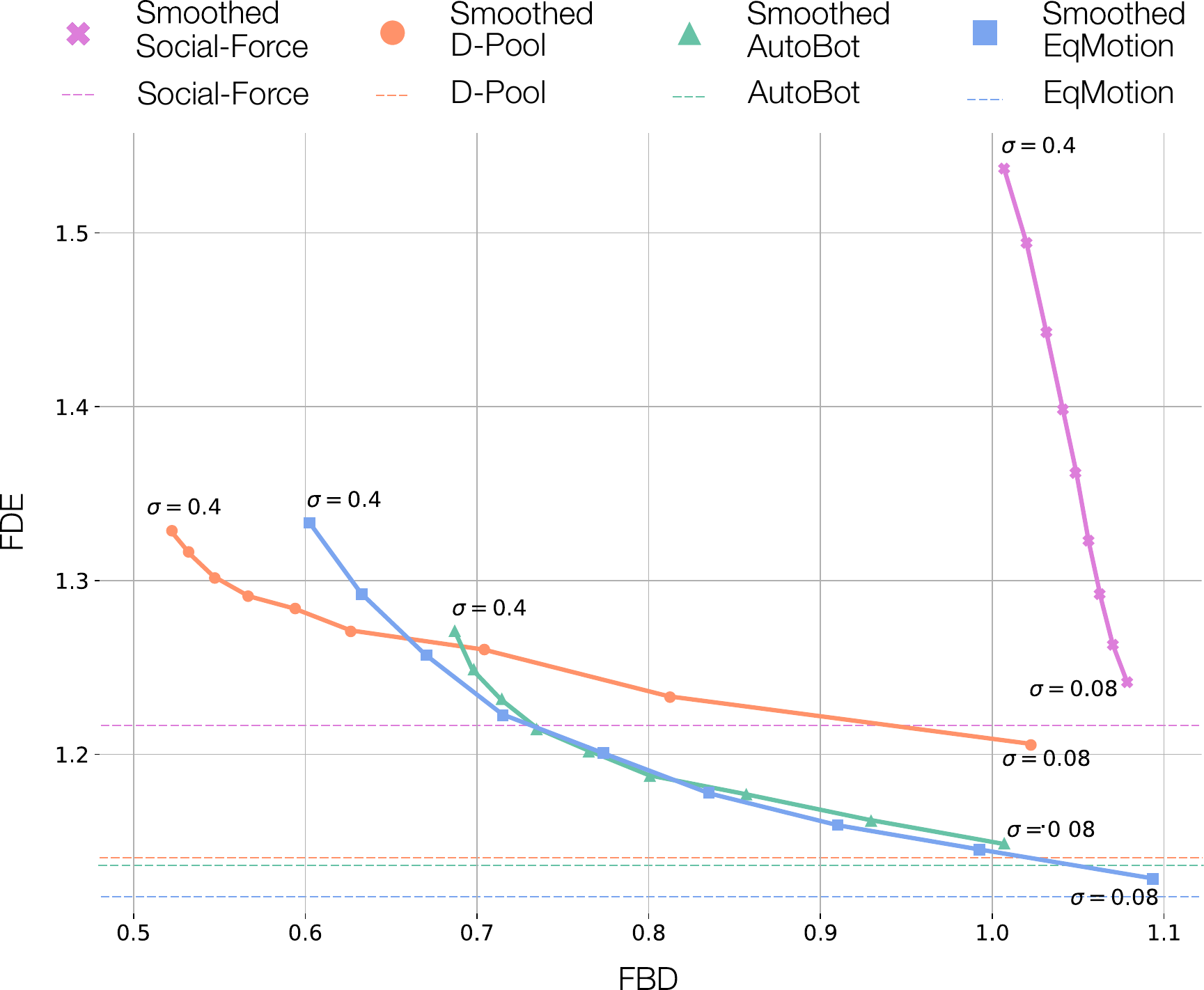}
    \includegraphics[width=\columnwidth]{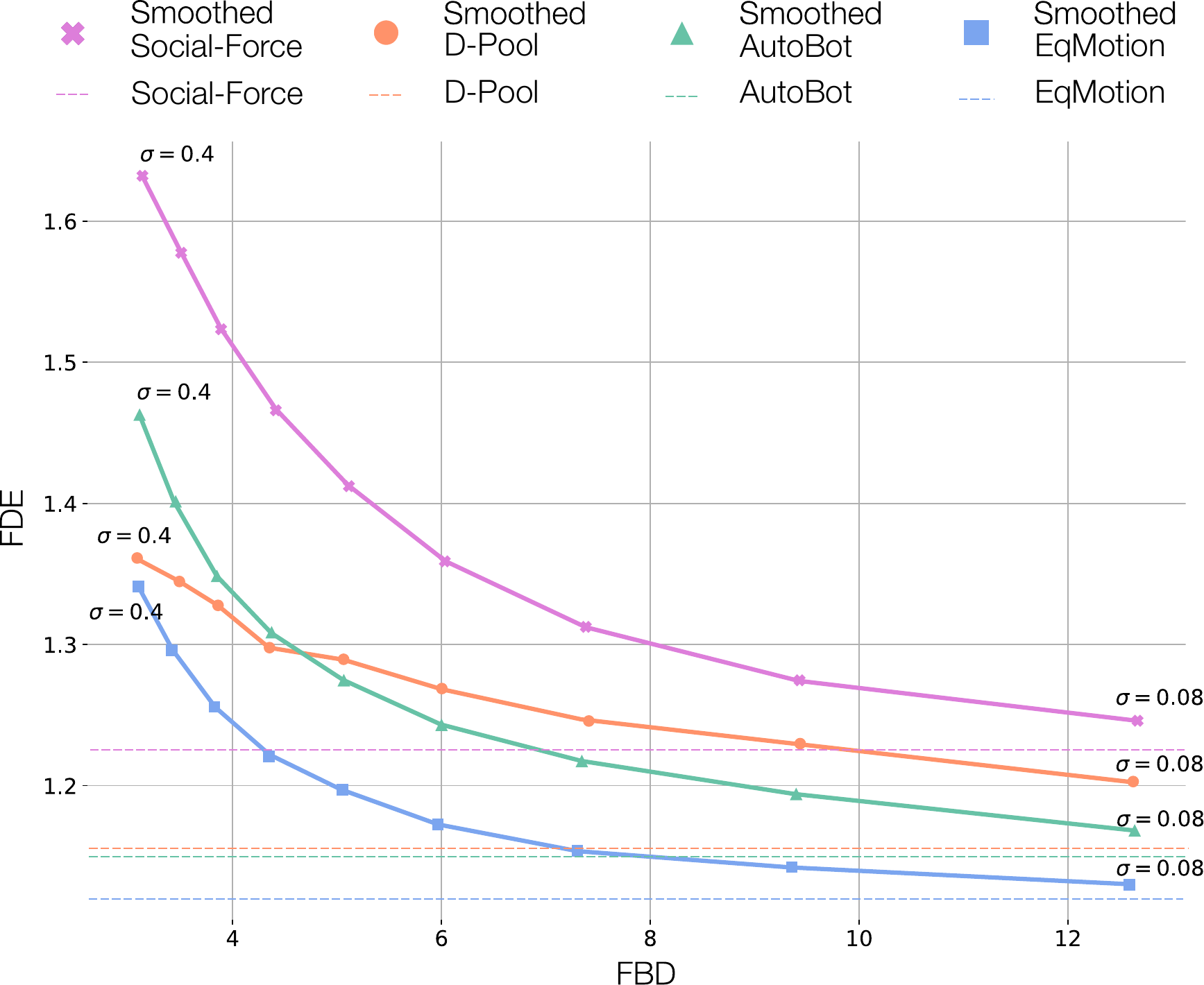}
    \caption{FDE against FBD for \textit{median} aggregation on the left and \textit{mean} aggregation on the right. The results are for different smoothed predictors with two aggregation functions and equally spaced $\sigma$ within $[0.08,0.4]$.
    The bottom left indicates the best performance. The figures show a trade-off between accuracy (represented in FDE) and robustness (represented as FBD). They also provide a comparison between models' guaranteed robustness.}
  \label{fig:median_baselines}
\end{figure*}

We initially report the performance of the predictors and their smoothed counterparts utilizing the median aggregation function in the left part of  \Cref{fig:median_baselines}. 
The figure shows the accuracy of models against the certified bounds, highlighting their accuracy with respect to robustness. Each point on the curves represents an instance of a smoothed predictor with a different hyperparameter $\sigma$. 
Therefore, changing $\sigma$ allows selecting a model instance with the desired trade off between accuracy and robustness.
While the original models (dashed lines in the figure) lack guarantees, the smoothed predictors provide certified bounds, albeit at the expense of a modest increase in FDE (ranging from $1\%$ to $6\%$ for different predictors with the smallest $\sigma$).
The figure also provides a way to compare the guaranteed robustness of different predictors. Given an FDE value, predictors with a smaller bound (\textit{i.e., }smaller FBD) have better guaranteed robustness. The figures show that while smoothed EqMotion and smoothed Autobot have similar bounds, smoothed D-Pool has a smaller bound for FDEs below $1.25$. 
Moreover, smoothed Social-Force has the largest bounds due to its significant sensitivity to input perturbations.

In the right part of \Cref{fig:median_baselines}, we show similar curves but using the mean aggregation function. Comparing the two sub-figures reveals that the median aggregator yields considerably smaller bounds compared to the mean aggregator, demonstrating the better alignment of the median with the trajectory prediction task.
This is probably because trajectory predictors are sensitive to input noise, leading to diverse outputs. Consequently, mean aggregation is more susceptible to outliers, whereas the median is less affected. Therefore, for our subsequent experiments, we opt for median aggregation. Moreover, we have selected EqMotion as our main predictor due its superior performance. In the rest of this section, we report experimental results to answer the remaining research questions.

\noindent\textbf{Is there a trade-off between accuracy and certified bound?}
As shown by \Cref{fig:median_baselines}, by increasing $\sigma$, the bounds progressively tighten while the accuracy drops, indicating a trade-off between them (see \Cref{eq:bounds1,eq:bounds2}).
Note that the hyperparameter $\sigma$ allows users to tailor the certified bound according to their needs.
For instance, given a desired FBD of $0.72$, we can choose $\sigma=0.28$ for smoothed EqMotion.

\noindent\textbf{Does the most accurate model have the best guaranteed robustness?} 
We report the performance of models with both certified and non-certified metrics in \Cref{tab:certified}. 
It shows that there is a large gap between FDE and Certified-FDE for all models, revealing models' lack of robustness.  
This shows the danger of solely relying on non-certified metrics. 
This analysis also uncovers a noteworthy observation: the model with the minimal FDE (EqMotion) is not the same as the model achieving the lowest Certified-FDE (D-Pool), indicating that a more accurate model is not necessarily more robust. 
It similarly shows a large gap between Col and Certified-Col for all models. 
As expected, Social-Force has the lowest collision rate due to its imposed collision-avoidance rules. However, the gap between Col and Certified-Col shows the model's sensitivity to input noise.

\begin{figure*}[!t]
\centering
    \centering
    \includegraphics[width=1.99\columnwidth]{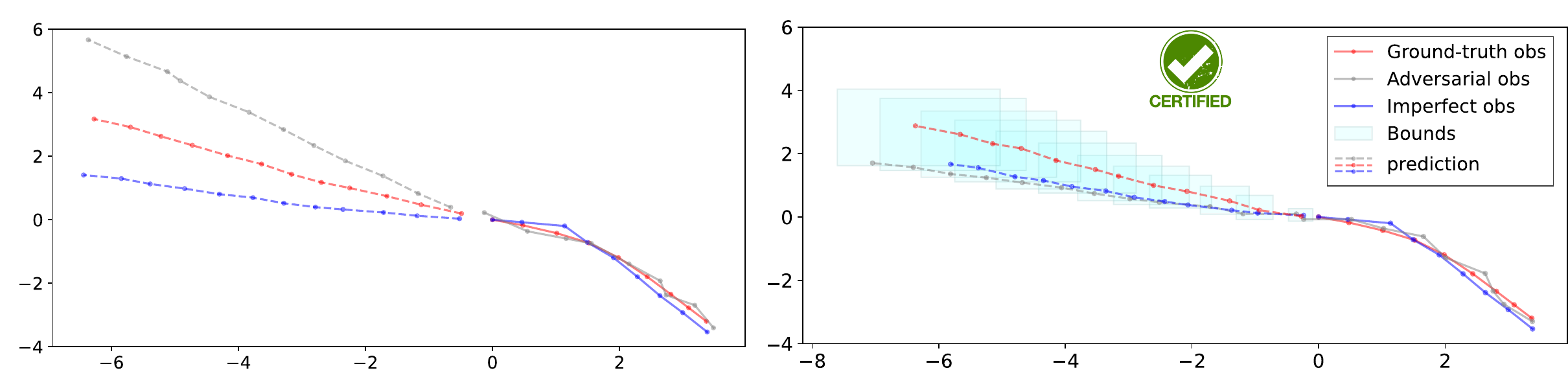}
\caption{Comparing the performance of the original (on the left) and the smoothed predictor (on the right). 
The original predictor's outputs change drastically with adversarial and imperfect inputs. In contrast, the smoothed predictor consistently predicts within the bounds, demonstrating higher reliability. For easier comparison, the final predicted points of the original predictor for the adversarial and imperfect observations are marked on the right figure with gray and blue stars, respectively.}
\label{fig:qual2}
\end{figure*}

\begin{table*}[!t]
  \centering
  \begin{tabular}{lccccc}
    \toprule
   Model & \;FDE\; & \;Certified-FDE\; & \;Col\; & \;Certified-Col\; \\
    \midrule
    Social-Force \cite{helbing1998social-forces} & 1.25 & N/A & 7.4 & N/A \\ 
    Smoothed Social-Force & 1.26 & 2.27 &8.0 & 46 \\
    \midrule
    D-Pool \cite{kothari2021human} & 1.14 & N/A & 9.4 & N/A\\
    Smoothed D-Pool & 1.23 & 2.0 & 9.0 & 49 \\
    \midrule
    AutoBot \cite{girgis2022autobot} & 1.14 & N/A & 8.8 & N/A\\
    Smoothed AutoBot & 1.17 & 2.05 & 9.3 & 53 \\
    \midrule
    EqMotion \cite{xu2023eqmotion} & 1.12 & N/A & 10.1 & N/A\\
    Smoothed EqMotion & 1.14 & 2.07 & 10.6 & 57\\
    \bottomrule
  \end{tabular}
    \caption{Comparing performances in terms of certified and non-certified metrics. Since non-smoothed models do not have any guarantee on their outputs, the certified metrics are not applicable (N/A) for them. 
  }
  \label{tab:certified}
\end{table*}

\noindent\textbf{What practical advantages does the smoothed predictor offer compared to the original model?}
To study this question, we examine the robustness of the models in two scenarios: adversarial attacks and real-world perception noise. 
We first investigate the robustness of the models against adversarial attacks by performing PGD attacks~\cite{madry2017pgd}. 
We demonstrate a scenario in \Cref{fig:qual2} where the left figure shows the existence of an adversarially perturbed input for the original model, leading to large deviations from the original prediction (more than 2m).
However, conducting PGD attacks on the smoothed predictor (the right figure) does not lead to predictions outside the certified bounds, demonstrating its guaranteed robustness. We provide a detailed quantitative analysis in the supplementary.  

In order to illustrate the impact of imperfect input data (\textit{i.e.,} noisy inputs) coming from perception systems on the predictors, we employ an off-the-shelf joint detection and tracking model~\cite{yin2021detection_tracking} to extract observation trajectories on the nuScenes dataset~\cite{Caesar2020nuScenesAM}. 
\Cref{fig:qual2} visualizes a real-world scenario with both the extracted imperfect observation sequence and the ground-truth. On the left, we observe that the imperfect observation influences the prediction of the model, leading to a large deviation from the prediction given ground-truth observation. This clearly shows that the performance of the model is sensitive to the input noise, making the model unreliable for safety-critical applications. In contrast, on the right, the predictions of the smoothed predictor for both observation sequences remain within the certified bounds, providing a reliable model.

\subsection{Discussions}
\label{sec:discussions}

\noindent \textbf{Denoiser analysis.}
Our proposed trajectory denoiser acts as a pre-processing module and therefore, incorporating it in the smoothing operation improves the certified bounds. In this part, we first compare the performance of different denoisers for trajectory denoising without considering the predictors. 
To this end, we measure the magnitude of the remaining noise in their outputs when provided with noisy input trajectories at different levels. 
In \Cref{tab:denoiser_snr_perf}, we report the performance of our proposed diffusion denoiser along with three established denoising methods for time-series data: the Wiener filter~\cite{wiener1949extrapolation} as a statistical approach, a Moving Average filter~\cite{rabiner1975theory} to filter high-frequency noise, and fitting a 4\textsuperscript{th} order polynomial that has been used previously to represent human trajectories~\cite{becker2018evaluation}. The results demonstrate that the diffusion denoiser outperforms other approaches in noise reduction.
Second, we evaluate the effect of the denoiser in the certified bounds of our smoothed predictor and compare it with not having a denoiser (assigning $h=\mathrm{id}$) in~\Cref{tab:denoising}. For similar FDE values, the smoothed predictor with diffusion denoiser has a significantly smaller certified bound, demonstrating the effectiveness of the denoiser in tightening the bounds.

\begin{table}[!t]
  \centering
  \small
    \centering
      \begin{tabular}{lrcc}
        \toprule
       Model & Noise = 0.08\ & 0.24\ & 0.40\ \\
        \midrule
        No denoiser & 0.08 & 0.24 & 0.40\\
        Polynomial & 0.08 & 0.22 & 0.36\\
        Moving Average & 0.07 & 0.18 & 0.29\\
        Wiener Filter & 0.06 & 0.16& 0.26\\
        Diffusion Denoiser (ours) & 0.06 & 0.14 & 0.24 \\
        \bottomrule
      \end{tabular}
\caption{Performance comparison of different denoisers. Noisy trajectories at three noise levels are fed to denoisers, and the remaining noise magnitude is reported.}
\label{tab:denoiser_snr_perf}
\end{table}

\begin{table}[!t]
\small
\centering
      \begin{tabular}{lrcc}
        \toprule
        Model & FDE = 1.2\; & 1.3\; & 1.4 \\
        \midrule
        W/o denoiser & 1.20 & 0.96 & 0.80\\
        W/ denoiser  & 0.78 & 0.65& 0.57\\
        \bottomrule
      \end{tabular}
\caption{Comparing FBD of the smoothed predictor with and without the denoiser across different FDE values.}
\label{tab:denoising}
\end{table}

\noindent\textbf{Multi-modal settings.}
To analyze certification in multi-modal settings, we employed the multi-modal EqMotion with $k=20$~\cite{xu2023eqmotion}. We report the multi-modal metrics defined in \Cref{sec:certrap} in~\Cref{tab:multimodal}. The multi-modal model is more accurate than its single-modal counterpart in terms of both FDE and Certified-FDE, as it captures diverse output modes. Moreover, it has a smaller FBD since each mode concentrates on a specific behavior, leading to a smaller certified bound.

\noindent\textbf{Multi-agent settings.}
In the real world, the observed trajectories of all agents can be noisy. Our method can be extended to this setting, where instead of perturbing the trajectory of one agent, we consider perturbations added to all agents in the scene.
\Cref{tab:multiagent} shows that with a similar FDE, the multi-agent model has a larger bound. Basically, when perturbing all agents, the interdependencies between agents make prediction change more, leading to larger certified bounds.

\begin{table}[!t]
\small
  \centering
    \centering
  \begin{tabular}{lccc}
    \toprule
    Model & FDE & FBD & Certified-FDE\\
    \midrule
    Single-modal  & 1.13 & 0.99 & 2.07 \\
    Multi-modal  & 0.39 & 0.64 & 1.38 \\
    \bottomrule
  \end{tabular}
\caption{Comparison of single- vs. multi-\textbf{modal} settings for Smoothed EqMotion.}
\label{tab:multimodal}
\end{table}

\begin{table}[!t]
\small
    \centering
  \begin{tabular}{lcc}
    \toprule
    Model & FDE & FBD \\
    \midrule
    Single-agent  & 1.13 & 0.99 \\
    Multi-agent  & 1.13 & 1.21 \\
    \bottomrule
  \end{tabular}
\label{tab:multiagent}
\caption{Comparison of certified bounds in single- vs. multi-\textbf{agent} settings in similar FDE values.}
\end{table}

\noindent\textbf{Downstream task.}
We also investigated whether improving trajectory prediction with the diffusion denoiser can enhance performance in a downstream task.
We considered the task of robot navigation in a dense crowd scenario employing a crowd navigation simulator~\cite{chen2019crowd}, where the objective for the robot is to navigate through a group of simulated pedestrians and reach a destination. At each timestep, the robot 
predicts the interactions, and determines its next action. 
We generated a dataset of 5,000 simulation episodes using the pre-trained SARL policy as the expert, and subsequently trained an imitation model on this data for 200 epochs~\cite{chen2019crowd}.
We adhered to their protocol, which measures the effectiveness of a policy using the collision rate, and accumulated reward. 

The results, summarized in \Cref{tab:sarl}, indicate that the learned policy is sensitive to input noise, which could be introduced in real-world scenarios. However, when our diffusion denoiser is incorporated, there is a consistent improvement in accumulated reward and a reduction in collision rate. This is due to the improved interaction prediction, resulting in better planning for the robot.

\begin{table}[!t]
\small
\centering
    \begin{tabular}{cc|cc}
    \toprule
    Method & Noise size & Reward {$\uparrow$} & Collision {(\%) $\downarrow$}\\ \midrule
    Original & 0 & 0.272 &  13.1\\
    Original & 0.2 & 0.230  & 21.0\\
    Robust & 0.2 & 0.263  & 15.1\\
    \bottomrule
    \end{tabular}
\caption{Quantitative results of crowd robot navigation~\cite{chen2019crowd} with different prediction methods. 
}
\label{tab:sarl}
\end{table}

\subsection{Limitations}

Randomized smoothing inevitably increases computational costs in order to provide guaranteed robustness. This is due to the fact that it requires evaluating the predictor $n$ times to obtain Monte-Carlo samples. Nonetheless, this process can be parallelized. 
Using an NVIDIA GeForce RTX 3090, the evaluation time for predicting a trajectory of $4.8$ seconds length in EqMotion is $0.07$ seconds, while for smoothed EqMotion with $n=100$, it is $0.1$ seconds which is small enough for many real-world  applications (only about $42\%$ increase in the computational time). As future work, one can improve it \textit{e.g.,} by better sampling strategies or optimized code structures.

\section{Conclusions}
\label{sec:conc}
In this work, we introduced a certified trajectory prediction approach that tackles the issue of lacking guaranteed robustness in the human trajectory prediction task. We also proposed a denoiser for trajectory data and introduced new certified metrics that ensure reliable performance assessments under noisy conditions. 
Throughout extensive experiments on various trajectory predictors, we found that the model with the highest accuracy is not always the most robust one. By adjusting certification parameters, one can prioritize either a tighter certified bound or higher accuracy. Moreover, our experiments demonstrated the edge of smoothed predictors over standard models in the presence of adversarial perturbations or input noise. We hope our work paves the way for more reliable trajectory predictors.

\section*{Acknowledgments}
The authors would like to thank Brian Sifringer, Taylor Mordan, Ahmad Rahimi, and Yuejiang Liu for their helpful comments. 
This project was partially funded by Honda R\&D Co., Ltd.

\clearpage


{
\small
\bibliographystyle{ieeenat_fullname}
\bibliography{ref}

\begin{thebibliography}{69}
\providecommand{\natexlab}[1]{#1}
\providecommand{\url}[1]{\texttt{#1}}
\expandafter\ifx\csname urlstyle\endcsname\relax
  \providecommand{\doi}[1]{doi: #1}\else
  \providecommand{\doi}{doi: \begingroup \urlstyle{rm}\Url}\fi

\bibitem[Alahi et~al.(2016)Alahi, Goel, Ramanathan, Robicquet, Fei-Fei, and
  Savarese]{alahi2016sociallstm}
Alexandre Alahi, Kratarth Goel, Vignesh Ramanathan, Alexandre Robicquet, Li
  Fei-Fei, and Silvio Savarese.
\newblock Social lstm: Human trajectory prediction in crowded spaces.
\newblock In \emph{Proceedings of the IEEE/CVF conference on Computer Vision
  and Pattern Recognition (CVPR)}, 2016.

\bibitem[Athalye et~al.(2018)Athalye, Carlini, and
  Wagner]{athalye2018obfuscated}
Anish Athalye, Nicholas Carlini, and David Wagner.
\newblock Obfuscated gradients give a false sense of security: Circumventing
  defenses to adversarial examples.
\newblock In \emph{International Conference on Machine Learning (ICML)}. PMLR,
  2018.

\bibitem[Bae et~al.(2024)Bae, Park, and Jeon]{bae2024singulartrajectory}
Inhwan Bae, Young-Jae Park, and Hae-Gon Jeon.
\newblock Singulartrajectory: Universal trajectory predictor using diffusion
  model.
\newblock In \emph{Proceedings of the IEEE/CVF Conference on Computer Vision
  and Pattern Recognition}, 2024.

\bibitem[Bahari et~al.(2022)Bahari, Saadatnejad, Rahimi, Shaverdikondori,
  Shahidzadeh, Moosavi-Dezfooli, and Alahi]{bahari2022sattack}
Mohammadhossein Bahari, Saeed Saadatnejad, Ahmad Rahimi, Mohammad
  Shaverdikondori, Amir-Hossein Shahidzadeh, Seyed-Mohsen Moosavi-Dezfooli, and
  Alexandre Alahi.
\newblock Vehicle trajectory prediction works, but not everywhere.
\newblock In \emph{Proceedings of the IEEE/CVF Conference on Computer Vision
  and Pattern Recognition (CVPR)}, 2022.

\bibitem[Bartoli et~al.(2018)Bartoli, Lisanti, Ballan, and
  Del~Bimbo]{bartoli2018context}
Federico Bartoli, Giuseppe Lisanti, Lamberto Ballan, and Alberto Del~Bimbo.
\newblock Context-aware trajectory prediction.
\newblock In \emph{International Conference on Pattern Recognition (ICPR)}.
  IEEE, 2018.

\bibitem[Becker et~al.(2018)Becker, Hug, H{\"u}bner, and
  Arens]{becker2018evaluation}
Stefan Becker, Ronny Hug, Wolfgang H{\"u}bner, and Michael Arens.
\newblock An evaluation of trajectory prediction approaches and notes on the
  trajnet benchmark.
\newblock \emph{arXiv preprint arXiv:1805.07663}, 2018.

\bibitem[Bunel et~al.(2018)Bunel, Turkaslan, Torr, Kohli, and
  Mudigonda]{bunel2018unified}
Rudy~R Bunel, Ilker Turkaslan, Philip Torr, Pushmeet Kohli, and Pawan~K
  Mudigonda.
\newblock A unified view of piecewise linear neural network verification.
\newblock \emph{Advances in Neural Information Processing Systems (NeurIPS)},
  31, 2018.

\bibitem[Caesar et~al.(2020)Caesar, Bankiti, Lang, Vora, Liong, Xu, Krishnan,
  Pan, Baldan, and Beijbom]{Caesar2020nuScenesAM}
Holger Caesar, Varun Bankiti, Alex~H. Lang, Sourabh Vora, Venice~Erin Liong,
  Qiang Xu, Anush Krishnan, Yu Pan, Giancarlo Baldan, and Oscar Beijbom.
\newblock nuscenes: A multimodal dataset for autonomous driving.
\newblock \emph{Proceedings of IEEE/CVF Conference on Computer Vision and
  Pattern Recognition (CVPR)}, 2020.

\bibitem[Cao et~al.(2021)Cao, Li, Ma, and Tomizuka]{cao2021spectral}
Defu Cao, Jiachen Li, Hengbo Ma, and Masayoshi Tomizuka.
\newblock Spectral temporal graph neural network for trajectory prediction.
\newblock In \emph{IEEE International Conference on Robotics and Automation
  (ICRA)}. IEEE, 2021.

\bibitem[Cao and Gong(2017)]{cao2017mitigating}
Xiaoyu Cao and Neil~Zhenqiang Gong.
\newblock Mitigating evasion attacks to deep neural networks via region-based
  classification.
\newblock In \emph{Proceedings of the 33rd Annual Computer Security
  Applications Conference}, pages 278--287, 2017.

\bibitem[Cao et~al.(2022)Cao, Xiao, Anandkumar, Xu, and Pavone]{cao2022advdo}
Yulong Cao, Chaowei Xiao, Anima Anandkumar, Danfei Xu, and Marco Pavone.
\newblock Advdo: Realistic adversarial attacks for trajectory prediction.
\newblock In \emph{European Conference on Computer Vision (ECCV)}, pages
  36--52. Springer, 2022.

\bibitem[Cao et~al.(2023)Cao, Xu, Weng, Mao, Anandkumar, Xiao, and
  Pavone]{cao2023robust}
Yulong Cao, Danfei Xu, Xinshuo Weng, Zhuoqing Mao, Anima Anandkumar, Chaowei
  Xiao, and Marco Pavone.
\newblock Robust trajectory prediction against adversarial attacks.
\newblock In \emph{Conference on Robot Learning (CoRL)}, pages 128--137. PMLR,
  2023.

\bibitem[Carlini and Wagner(2017)]{carlini2017adversarial}
Nicholas Carlini and David Wagner.
\newblock Adversarial examples are not easily detected: Bypassing ten detection
  methods.
\newblock In \emph{Proceedings of the 10th ACM workshop on artificial
  intelligence and security}, pages 3--14, 2017.

\bibitem[Carlini et~al.(2022)Carlini, Tramer, Dvijotham, Rice, Sun, and
  Kolter]{carlini2022certified_diffusion}
Nicholas Carlini, Florian Tramer, Krishnamurthy~Dj Dvijotham, Leslie Rice,
  Mingjie Sun, and J~Zico Kolter.
\newblock (certified!!) adversarial robustness for free!
\newblock \emph{arXiv preprint arXiv:2206.10550}, 2022.

\bibitem[Chavdarova et~al.(2018)Chavdarova, Baqu{\'e}, Bouquet, Maksai, Jose,
  Bagautdinov, Lettry, Fua, Gool, and Fleuret]{chavdarova2018wildtrack}
Tatjana Chavdarova, Pierre Baqu{\'e}, St{\'e}phane Bouquet, Andrii Maksai, Cijo
  Jose, Timur~M. Bagautdinov, Louis Lettry, Pascal Fua, Luc~Van Gool, and
  François Fleuret.
\newblock Wildtrack: A multi-camera hd dataset for dense unscripted pedestrian
  detection.
\newblock \emph{Prooceedings of the IEEE/CVF Conference on Computer Vision and
  Pattern Recognition (CVPR)}, 2018.

\bibitem[Chen et~al.(2019)Chen, Liu, Kreiss, and Alahi]{chen2019crowd}
Changan Chen, Yuejiang Liu, Sven Kreiss, and Alexandre Alahi.
\newblock Crowd-{Robot} {Interaction}: {Crowd}-{Aware} {Robot} {Navigation}
  {With} {Attention}-{Based} {Deep} {Reinforcement} {Learning}.
\newblock In \emph{IEEE International Conference on Robotics and Automation
  (ICRA)}, pages 6015--6022, 2019.
\newblock ISSN: 2577-087X.

\bibitem[Chiang et~al.(2020)Chiang, Curry, Abdelkader, Kumar, Dickerson, and
  Goldstein]{chiang2020detection_as_regression}
Ping-yeh Chiang, Michael Curry, Ahmed Abdelkader, Aounon Kumar, John Dickerson,
  and Tom Goldstein.
\newblock Detection as regression: Certified object detection with median
  smoothing.
\newblock \emph{Advances in Neural Information Processing Systems (NeurIPS)},
  2020.

\bibitem[Cohen et~al.(2019)Cohen, Rosenfeld, and
  Kolter]{cohen2019certified_kolter}
Jeremy Cohen, Elan Rosenfeld, and Zico Kolter.
\newblock Certified adversarial robustness via randomized smoothing.
\newblock In \emph{International Conference on Machine Learning (ICML)}, pages
  1310--1320. PMLR, 2019.

\bibitem[Dvijotham et~al.(2018)Dvijotham, Stanforth, Gowal, Mann, and
  Kohli]{dvijotham2018dual}
Krishnamurthy Dvijotham, Robert Stanforth, Sven Gowal, Timothy~A Mann, and
  Pushmeet Kohli.
\newblock A dual approach to scalable verification of deep networks.
\newblock In \emph{UAI}, page~3, 2018.

\bibitem[Ehlers(2017)]{ehlers2017formal}
Ruediger Ehlers.
\newblock Formal verification of piece-wise linear feed-forward neural
  networks.
\newblock In \emph{Automated Technology for Verification and Analysis: 15th
  International Symposium, ATVA 2017, Pune, India, October 3--6, 2017,
  Proceedings 15}, pages 269--286. Springer, 2017.

\bibitem[Fischetti and Jo(2018)]{fischetti2018deep}
Matteo Fischetti and Jason Jo.
\newblock Deep neural networks and mixed integer linear optimization.
\newblock \emph{Constraints}, 23\penalty0 (3):\penalty0 296--309, 2018.

\bibitem[Franco et~al.(2023)Franco, Placidi, Giuliari, Hasan, Cristani, and
  Galasso]{franco2023under}
Luca Franco, Leonardo Placidi, Francesco Giuliari, Irtiza Hasan, Marco
  Cristani, and Fabio Galasso.
\newblock Under the hood of transformer networks for trajectory forecasting.
\newblock \emph{Pattern Recognition}, 138:\penalty0 109372, 2023.

\bibitem[Gal et~al.(2022)Gal, Alaluf, Atzmon, Patashnik, Bermano, Chechik, and
  Cohen-Or]{gal2022image}
Rinon Gal, Yuval Alaluf, Yuval Atzmon, Or Patashnik, Amit~H Bermano, Gal
  Chechik, and Daniel Cohen-Or.
\newblock An image is worth one word: Personalizing text-to-image generation
  using textual inversion.
\newblock \emph{arXiv preprint arXiv:2208.01618}, 2022.

\bibitem[Girgis et~al.(2022)Girgis, Golemo, Codevilla, Weiss, D'Souza, Kahou,
  Heide, and Pal]{girgis2022autobot}
Roger Girgis, Florian Golemo, Felipe Codevilla, Martin Weiss, Jim~Aldon
  D'Souza, Samira~Ebrahimi Kahou, Felix Heide, and Christopher Pal.
\newblock Latent variable sequential set transformers for joint multi-agent
  motion prediction.
\newblock In \emph{International Conference on Learning Representations
  (ICLR)}, 2022.

\bibitem[Giuliari et~al.(2021)Giuliari, Hasan, Cristani, and
  Galasso]{giuliari2021transformer}
Francesco Giuliari, Irtiza Hasan, Marco Cristani, and Fabio Galasso.
\newblock Transformer networks for trajectory forecasting.
\newblock In \emph{2020 25th international conference on pattern recognition
  (ICPR)}. IEEE, 2021.

\bibitem[Gu et~al.(2022)Gu, Chen, Li, Lin, Rao, Zhou, and Lu]{gu2022stochastic}
Tianpei Gu, Guangyi Chen, Junlong Li, Chunze Lin, Yongming Rao, Jie Zhou, and
  Jiwen Lu.
\newblock Stochastic trajectory prediction via motion indeterminacy diffusion.
\newblock In \emph{Proceedings of the IEEE/CVF Conference on Computer Vision
  and Pattern Recognition (CVPR)}, pages 17113--17122, 2022.

\bibitem[Helbing and Molnar(1998)]{helbing1998social-forces}
Dirk Helbing and Peter Molnar.
\newblock Social force model for pedestrian dynamics.
\newblock \emph{Physical Review E}, 51, 1998.

\bibitem[Ho et~al.(2020)Ho, Jain, and Abbeel]{ho2020ddpm}
Jonathan Ho, Ajay Jain, and Pieter Abbeel.
\newblock Denoising diffusion probabilistic models.
\newblock \emph{arXiv preprint arxiv:2006.11239}, 2020.

\bibitem[Huang et~al.(2017)Huang, Kwiatkowska, Wang, and Wu]{huang2017safety}
Xiaowei Huang, Marta Kwiatkowska, Sen Wang, and Min Wu.
\newblock Safety verification of deep neural networks.
\newblock In \emph{Computer Aided Verification: 29th International Conference,
  CAV 2017, Heidelberg, Germany, July 24-28, 2017, Proceedings, Part I 30},
  pages 3--29. Springer, 2017.

\bibitem[H{\"u}llermeier and Waegeman(2021)]{hullermeier2021aleatoric}
Eyke H{\"u}llermeier and Willem Waegeman.
\newblock Aleatoric and epistemic uncertainty in machine learning: An
  introduction to concepts and methods.
\newblock \emph{Machine Learning}, 110:\penalty0 457--506, 2021.

\bibitem[Ivanovic and Pavone(2019)]{ivanovic2019trajectron}
Boris Ivanovic and Marco Pavone.
\newblock The trajectron: Probabilistic multi-agent trajectory modeling with
  dynamic spatiotemporal graphs.
\newblock In \emph{Proceedings of the IEEE/CVF International Conference on
  Computer Vision (ICCV)}, 2019.

\bibitem[Jiao et~al.(2023)Jiao, Liu, Sato, Chen, and Zhu]{jiao2023semi}
Ruochen Jiao, Xiangguo Liu, Takami Sato, Qi~Alfred Chen, and Qi Zhu.
\newblock Semi-supervised semantics-guided adversarial training for robust
  trajectory prediction.
\newblock In \emph{Proceedings of the IEEE/CVF International Conference on
  Computer Vision (ICCV)}, pages 8207--8217, 2023.

\bibitem[Khachatryan et~al.(2023)Khachatryan, Movsisyan, Tadevosyan, Henschel,
  Wang, Navasardyan, and Shi]{khachatryan2023text2video}
Levon Khachatryan, Andranik Movsisyan, Vahram Tadevosyan, Roberto Henschel,
  Zhangyang Wang, Shant Navasardyan, and Humphrey Shi.
\newblock Text2video-zero: Text-to-image diffusion models are zero-shot video
  generators.
\newblock In \emph{Proceedings of the IEEE/CVF International Conference on
  Computer Vision (ICCV)}, 2023.

\bibitem[Kothari et~al.(2021{\natexlab{a}})Kothari, Kreiss, and
  Alahi]{kothari2021human}
Parth Kothari, Sven Kreiss, and Alexandre Alahi.
\newblock Human trajectory forecasting in crowds: A deep learning perspective.
\newblock \emph{IEEE Transactions on Intelligent Transportation Systems},
  2021{\natexlab{a}}.

\bibitem[Kothari et~al.(2021{\natexlab{b}})Kothari, Sifringer, and
  Alahi]{kothari2021interpretable}
Parth Kothari, Brian Sifringer, and Alexandre Alahi.
\newblock Interpretable social anchors for human trajectory forecasting in
  crowds.
\newblock In \emph{Proceedings of the IEEE/CVF Conference on Computer Vision
  and Pattern Recognition (CVPR)}, 2021{\natexlab{b}}.

\bibitem[Lee et~al.(2024)Lee, Lee, Yu, Kim, and Lee]{lee2024mart}
Seongju Lee, Junseok Lee, Yeonguk Yu, Taeri Kim, and Kyoobin Lee.
\newblock Mart: Multiscale relational transformer networks for multi-agent
  trajectory prediction.
\newblock In \emph{European Conference on Computer Vision (ECCV)}. Springer,
  2024.

\bibitem[Lerner et~al.(2007)Lerner, Chrysanthou, and Lischinski]{lerner2007ucy}
Alon Lerner, Yiorgos Chrysanthou, and Dani Lischinski.
\newblock Crowds by example.
\newblock \emph{Comput. Graph. Forum}, 26, 2007.

\bibitem[Liu et~al.(2018)Liu, Cheng, Zhang, and Hsieh]{liu2018towards}
Xuanqing Liu, Minhao Cheng, Huan Zhang, and Cho-Jui Hsieh.
\newblock Towards robust neural networks via random self-ensemble.
\newblock In \emph{European Conference on Computer Vision (ECCV)}, 2018.

\bibitem[Liu et~al.(2021)Liu, Yan, and Alahi]{liu2021socialnce}
Yuejiang Liu, Qi Yan, and Alexandre Alahi.
\newblock Social nce: Contrastive learning of socially-aware motion
  representations.
\newblock In \emph{Proceedings of the IEEE/CVF International Conference on
  Computer Vision (ICCV)}, 2021.

\bibitem[Madry et~al.(2017{\natexlab{a}})Madry, Makelov, Schmidt, Tsipras, and
  Vladu]{madry2017pgd}
Aleksander Madry, Aleksandar Makelov, Ludwig Schmidt, Dimitris Tsipras, and
  Adrian Vladu.
\newblock Towards deep learning models resistant to adversarial attacks.
\newblock \emph{arXiv preprint arXiv:1706.06083}, 2017{\natexlab{a}}.

\bibitem[Madry et~al.(2017{\natexlab{b}})Madry, Makelov, Schmidt, Tsipras, and
  Vladu]{madry2017pgdattack}
Aleksander Madry, Aleksandar Makelov, Ludwig Schmidt, Dimitris Tsipras, and
  Adrian Vladu.
\newblock Towards deep learning models resistant to adversarial attacks.
\newblock \emph{arXiv preprint arXiv:1706.06083}, 2017{\natexlab{b}}.

\bibitem[Mao et~al.(2023)Mao, Xu, Zhu, Chen, and Wang]{mao2023leapfrog}
Weibo Mao, Chenxin Xu, Qi Zhu, Siheng Chen, and Yanfeng Wang.
\newblock Leapfrog diffusion model for stochastic trajectory prediction.
\newblock In \emph{Proceedings of the IEEE/CVF Conference on Computer Vision
  and Pattern Recognition}, pages 5517--5526, 2023.

\bibitem[Mohamed et~al.(2020)Mohamed, Qian, Elhoseiny, and
  Claudel]{mohamed2020social}
Abduallah Mohamed, Kun Qian, Mohamed Elhoseiny, and Christian Claudel.
\newblock Social-stgcnn: A social spatio-temporal graph convolutional neural
  network for human trajectory prediction.
\newblock In \emph{Proceedings of the IEEE/CVF Conference on Computer Vision
  and Pattern Recognition (CVPR)}, pages 14424--14432, 2020.

\bibitem[Nikhil and Tran~Morris(2018)]{nikhil2018convolutional}
Nishant Nikhil and Brendan Tran~Morris.
\newblock Convolutional neural network for trajectory prediction.
\newblock In \emph{European Conference on Computer Vision (ECCV) Workshops},
  2018.

\bibitem[Pellegrini et~al.(2010)Pellegrini, Ess, and Gool]{pellegrini2010eth}
Stefano Pellegrini, Andreas Ess, and Luc~Van Gool.
\newblock Improving data association by joint modeling of pedestrian
  trajectories and groupings.
\newblock In \emph{European Conference on Computer Vision (ECCV)}. Springer,
  2010.

\bibitem[Rabiner and Gold(1975)]{rabiner1975theory}
Lawrence~R Rabiner and Bernard Gold.
\newblock Theory and application of digital signal processing.
\newblock \emph{Englewood Cliffs: Prentice-Hall}, 1975.

\bibitem[Rombach et~al.(2022)Rombach, Blattmann, Lorenz, Esser, and
  Ommer]{rombach2022high}
Robin Rombach, Andreas Blattmann, Dominik Lorenz, Patrick Esser, and Bj{\"o}rn
  Ommer.
\newblock High-resolution image synthesis with latent diffusion models.
\newblock In \emph{Proceedings of the IEEE/CVF Conference on Computer Vision
  and Pattern Recognition (CVPR)}, pages 10684--10695, 2022.

\bibitem[Saadatnejad et~al.(2022)Saadatnejad, Bahari, Khorsandi, Saneian,
  Moosavi-Dezfooli, and Alahi]{saadatnejad2021sattack}
Saeed Saadatnejad, Mohammadhossein Bahari, Pedram Khorsandi, Mohammad Saneian,
  Seyed-Mohsen Moosavi-Dezfooli, and Alexandre Alahi.
\newblock Are socially-aware trajectory prediction models really
  socially-aware?
\newblock \emph{Transportation Research Part C: Emerging Technologies},
  141:\penalty0 103705, 2022.

\bibitem[Saadatnejad et~al.(2023)Saadatnejad, Rasekh, Mofayezi, Medghalchi,
  Rajabzadeh, Mordan, and Alahi]{saadatnejad2023diffusion}
Saeed Saadatnejad, Ali Rasekh, Mohammadreza Mofayezi, Yasamin Medghalchi, Sara
  Rajabzadeh, Taylor Mordan, and Alexandre Alahi.
\newblock A generic diffusion-based approach for 3d human pose prediction in
  the wild.
\newblock In \emph{International Conference on Robotics and Automation (ICRA)},
  2023.

\bibitem[Saadatnejad et~al.(2024{\natexlab{a}})Saadatnejad, Gao, Messaoud, and
  Alahi]{saadatnejad2024socialtransmotion}
Saeed Saadatnejad, Yang Gao, Kaouther Messaoud, and Alexandre Alahi.
\newblock Social-transmotion: Promptable human trajectory prediction.
\newblock In \emph{International Conference on Learning Representations
  (ICLR)}, 2024{\natexlab{a}}.

\bibitem[Saadatnejad et~al.(2024{\natexlab{b}})Saadatnejad, Mirmohammadi,
  Daghyani, Saremi, Benisi, Alimohammadi, Tehraninasab, Mordan, and
  Alahi]{saadatnejad2024toward}
Saeed Saadatnejad, Mehrshad Mirmohammadi, Matin Daghyani, Parham Saremi,
  Yashar~Zoroofchi Benisi, Amirhossein Alimohammadi, Zahra Tehraninasab, Taylor
  Mordan, and Alexandre Alahi.
\newblock Toward reliable human pose forecasting with uncertainty.
\newblock \emph{IEEE Robotics and Automation Letters (RA-L)},
  2024{\natexlab{b}}.

\bibitem[Salman et~al.(2020)Salman, Sun, Yang, Kapoor, and
  Kolter]{salman2020denoised}
Hadi Salman, Mingjie Sun, Greg Yang, Ashish Kapoor, and J~Zico Kolter.
\newblock Denoised smoothing: A provable defense for pretrained classifiers.
\newblock \emph{Advances in Neural Information Processing Systems (NeurIPS)},
  33:\penalty0 21945--21957, 2020.

\bibitem[Shafer and Vovk(2008)]{shafer2008tutorial}
Glenn Shafer and Vladimir Vovk.
\newblock A tutorial on conformal prediction.
\newblock \emph{Journal of Machine Learning Research}, 9\penalty0 (3), 2008.

\bibitem[Singh et~al.(2018)Singh, Gehr, Mirman, P{\"u}schel, and
  Vechev]{singh2018fast}
Gagandeep Singh, Timon Gehr, Matthew Mirman, Markus P{\"u}schel, and Martin
  Vechev.
\newblock Fast and effective robustness certification.
\newblock \emph{Advances in Neural Information Processing Systems (NeurIPS)},
  31, 2018.

\bibitem[Tan et~al.(2023)Tan, Wang, and Kantaros]{tan2023targeted}
Kaiyuan Tan, Jun Wang, and Yiannis Kantaros.
\newblock Targeted adversarial attacks against neural network trajectory
  predictors.
\newblock In \emph{Learning for Dynamics and Control Conference}, pages
  431--444. PMLR, 2023.

\bibitem[Uesato et~al.(2018)Uesato, O’donoghue, Kohli, and
  Oord]{uesato2018adversarial}
Jonathan Uesato, Brendan O’donoghue, Pushmeet Kohli, and Aaron Oord.
\newblock Adversarial risk and the dangers of evaluating against weak attacks.
\newblock In \emph{International Conference on Machine Learning (ICML)}. PMLR,
  2018.

\bibitem[Wang et~al.(2024)Wang, Tang, Sun, Rossi, Xie, Peng, Hannagan,
  Sabatini, Poerio, Tomizuka, et~al.]{wang2024optimizing}
Yixiao Wang, Chen Tang, Lingfeng Sun, Simone Rossi, Yichen Xie, Chensheng Peng,
  Thomas Hannagan, Stefano Sabatini, Nicola Poerio, Masayoshi Tomizuka, et~al.
\newblock Optimizing diffusion models for joint trajectory prediction and
  controllable generation.
\newblock In \emph{European Conference on Computer Vision (ECCV)}. Springer,
  2024.

\bibitem[Wiener(1949)]{wiener1949extrapolation}
Norbert Wiener.
\newblock \emph{Extrapolation, interpolation, and smoothing of stationary time
  series: with engineering applications}.
\newblock The MIT press, 1949.

\bibitem[Wong and Kolter(2018)]{wong2018provable}
Eric Wong and Zico Kolter.
\newblock Provable defenses against adversarial examples via the convex outer
  adversarial polytope.
\newblock In \emph{International Conference on Machine Learning (ICML)}, pages
  5286--5295. PMLR, 2018.

\bibitem[Xu et~al.(2023)Xu, Tan, Tan, Chen, Wang, Wang, and
  Wang]{xu2023eqmotion}
Chenxin Xu, Robby~T Tan, Yuhong Tan, Siheng Chen, Yu~Guang Wang, Xinchao Wang,
  and Yanfeng Wang.
\newblock Eqmotion: Equivariant multi-agent motion prediction with invariant
  interaction reasoning.
\newblock In \emph{Proceedings of the IEEE/CVF Conference on Computer Vision
  and Pattern Recognition (CVPR)}, 2023.

\bibitem[Xu et~al.(2024)Xu, Chambon, Chen, Alahi, Cord, Perez,
  et~al.]{xu2024towards}
Yihong Xu, Loick Chambon, Micka{\"e}l Chen, Alexandre Alahi, Matthieu Cord,
  Patrick Perez, et~al.
\newblock Towards motion forecasting with real-world perception inputs: Are
  end-to-end approaches competitive?
\newblock In \emph{International Conference on Robotics and Automation (ICRA)},
  2024.

\bibitem[Yang et~al.(2023)Yang, Liang, and Su]{yang2023real}
Cheng Yang, Lijing Liang, and Zhixun Su.
\newblock Real-world denoising via diffusion model.
\newblock \emph{arXiv preprint arXiv:2305.04457}, 2023.

\bibitem[Yin et~al.(2021)Yin, Zhou, and Krahenbuhl]{yin2021detection_tracking}
Tianwei Yin, Xingyi Zhou, and Philipp Krahenbuhl.
\newblock Center-based 3d object detection and tracking.
\newblock In \emph{Proceedings of the IEEE/CVF Conference on Computer Vision
  and Pattern Recognition (CVPR)}, pages 11784--11793, 2021.

\bibitem[Zamboni et~al.(2022)Zamboni, Kefato, Girdzijauskas, Nor{\'e}n, and
  Dal~Col]{zamboni2022pedestrian}
Simone Zamboni, Zekarias~Tilahun Kefato, Sarunas Girdzijauskas, Christoffer
  Nor{\'e}n, and Laura Dal~Col.
\newblock Pedestrian trajectory prediction with convolutional neural networks.
\newblock \emph{Pattern Recognition}, 121:\penalty0 108252, 2022.

\bibitem[Zhang et~al.(2023{\natexlab{a}})Zhang, Rao, and
  Agrawala]{zhang2023adding}
Lvmin Zhang, Anyi Rao, and Maneesh Agrawala.
\newblock Adding conditional control to text-to-image diffusion models.
\newblock In \emph{Proceedings of the IEEE/CVF International Conference on
  Computer Vision (ICCV)}, pages 3836--3847, 2023{\natexlab{a}}.

\bibitem[Zhang et~al.(2023{\natexlab{b}})Zhang, Xu, Yang, Jin, Huang, and
  Zhang]{zhang2023trajpac}
Liang Zhang, Nathaniel Xu, Pengfei Yang, Gaojie Jin, Cheng-Chao Huang, and
  Lijun Zhang.
\newblock Trajpac: Towards robustness verification of pedestrian trajectory
  prediction models.
\newblock In \emph{Proceedings of the IEEE/CVF International Conference on
  Computer Vision (ICCV)}, pages 8327--8339, 2023{\natexlab{b}}.

\bibitem[Zhang et~al.(2022)Zhang, Hu, Sun, Chen, and Mao]{zhang2022adversarial}
Qingzhao Zhang, Shengtuo Hu, Jiachen Sun, Qi~Alfred Chen, and Z~Morley Mao.
\newblock On adversarial robustness of trajectory prediction for autonomous
  vehicles.
\newblock In \emph{Proceedings of the IEEE/CVF Conference on Computer Vision
  and Pattern Recognition (CVPR)}, pages 15159--15168, 2022.

\bibitem[Zhu et~al.(2023{\natexlab{a}})Zhu, Ye, Zhang, Zhao, and
  Yu]{zhu2023difftraj}
Yuanshao Zhu, Yongchao Ye, Shiyao Zhang, Xiangyu Zhao, and James Yu.
\newblock Difftraj: Generating {GPS} trajectory with diffusion probabilistic
  model.
\newblock In \emph{Thirty-seventh Conference on Neural Information Processing
  Systems}, 2023{\natexlab{a}}.

\bibitem[Zhu et~al.(2023{\natexlab{b}})Zhu, Zhang, Liang, Cao, Wen, Timofte,
  and Van~Gool]{zhu2023denoising}
Yuanzhi Zhu, Kai Zhang, Jingyun Liang, Jiezhang Cao, Bihan Wen, Radu Timofte,
  and Luc Van~Gool.
\newblock Denoising diffusion models for plug-and-play image restoration.
\newblock In \emph{Proceedings of the IEEE/CVF Conference on Computer Vision
  and Pattern Recognition}, pages 1219--1229, 2023{\natexlab{b}}.

\end{thebibliography}
}

\clearpage
\setcounter{page}{1}
\maketitlesupplementary

Here, we provide additional content to complement our main paper. It includes more details of the method regarding the diffusion denoiser in \Cref{sec:diffusion}, and the algorithm  in \Cref{sec:code}. We then provide additional results complementing our experiments. Specifically, we provide results using other metrics in \Cref{sec:quantitative},
bring an expanded set of qualitative results in \Cref{sec:qualitative}, 
demonstrate the bound sizes per timestep in \Cref{sec:timestep} and analyze the bounds with adversarial attacks in \Cref{sec:adv_appendix}.

\section{Further details of the method}
Here, we provide more details about the methodology of the paper. We first elaborate on the diffusion denoiser, then provide an algorithm for the method. 
\subsection{Diffusion denoiser}
\label{sec:diffusion}

For the denoiser architecture, we used a simple model comprising four residual 1D convolution layers and two linear layers with ReLU activation functions and dropout. This model was then trained using the diffusion approach described below.

At training time, given each input data \( x^0 \), our diffusion model selects a step $t$ and then adds Gaussian noise with zero mean and a pre-defined variance to the input to create a noisier version \( x^1 \). This process is repeated for \( t \) steps resulting in a noisy \( x^t \):
\[
q(x^{t} | x^{t-1}) = x^{t-1} + \mathcal{N}(x^t; \sqrt{1-\beta^t} x^{t-1}, \beta^t \mathbf{I}),
\]
where \( q \) denotes the forward process, and \( \beta^t \) is the variance of the noise at step \( t \), determined by a scheduler. We utilize a linear noise scheduler in our denoiser. The network learns to reverse the diffusion process and recover the clean signal by predicting the cumulative noise added to \( x^t \).

At inference time, the model starts with a noisy input trajectory \( x^t \). The step \(t\) is estimated according to the given $\sigma$ and the scheduler, and then model iteratively predicts the less noisy signal\footnote{Note that rescaling factors are applied at each step to match the expected variance of the diffusion model.}, reducing the noise step-by-step to get $x^{t-1}, x^{t-2}, \cdots$ until it obtains \( x^0 \). This is achieved by subtracting the additive noise learned during training from the output of the previous step, ultimately recovering the original signal.

\subsection{Algorithm}
\label{sec:code}

\begin{algorithm}[!h]
\caption{Smoothed Trajectory Prediction and its Certified Bounds}
\begin{algorithmic}[1]
\State \textbf{Input:} Input trajectory $X$, number of Monte-Carlo samples $n$, number of predictions $k$, aggregation operator $\mathcal{A}$, trajectory predictor $g$, denoiser $h$, certification radius $R$, hyperparameter $\sigma$, lower bounds $\{l_j\}$, upper bounds $\{u_j\}$
\State \textbf{Output:} Certified trajectory prediction $\tilde{f}(X)$, the certified bounds
\Procedure{}{}
    \State Initialize an empty list $arr$ to store predictions
    \For{$i = 1$ to $n$}
        \State $\epsilon^i \sim \mathcal{N}(0, \sigma^2 I)$ \Comment{Acquire a sample from the Gaussian distribution}
        \State $X^i \gets X + \epsilon^i$ \Comment{Generate perturbed inputs}
        \State $f(X^i) = g(h(X^i))$ \Comment{Process through denoiser $h$ and predictor $g$}
        \If{$\mathcal{A}$==Mean} 
            \State Clamp the $j$-th coordinate of $f(.)$ within $[l_j, u_j]$ \Comment{Adaptive clamping}
        \EndIf
        \State Append $f^k(X^i)$ to $arr$ \Comment{Certify $k$ modes}
    \EndFor
    \State $Y \gets \mathcal{A}(arr)$ \Comment{Aggregate the predictions with point-wise mean or median}
    \If{$\mathcal{A}$ == Mean} \Comment{Bounds for mean}
        \State Compute LB and UB on $Y$ from \Cref{eq:bounds1}, 
        given $R, \{l_j\}, \{u_j\}$
    \Else \Comment{Bounds for median}
        \State Compute LB and UB on $Y$ from \Cref{eq:bounds2}, 
        given $R$
    \EndIf
    \State \textbf{return} $Y$, LB, UB \Comment{Return prediction and certified bounds}
\EndProcedure
\end{algorithmic}
\label{alg:main}
\end{algorithm}

\Cref{alg:main} provides a high-level overview of our method.
For an explanation of the notation used, refer to \Cref{sec:method} of the paper.

\section{Additional results}
\subsection{Results using ADE, ABD and Certified-ADE}
\label{sec:quantitative}
\begin{figure*}[!t]
    \centering
    \includegraphics[width=0.99\columnwidth]{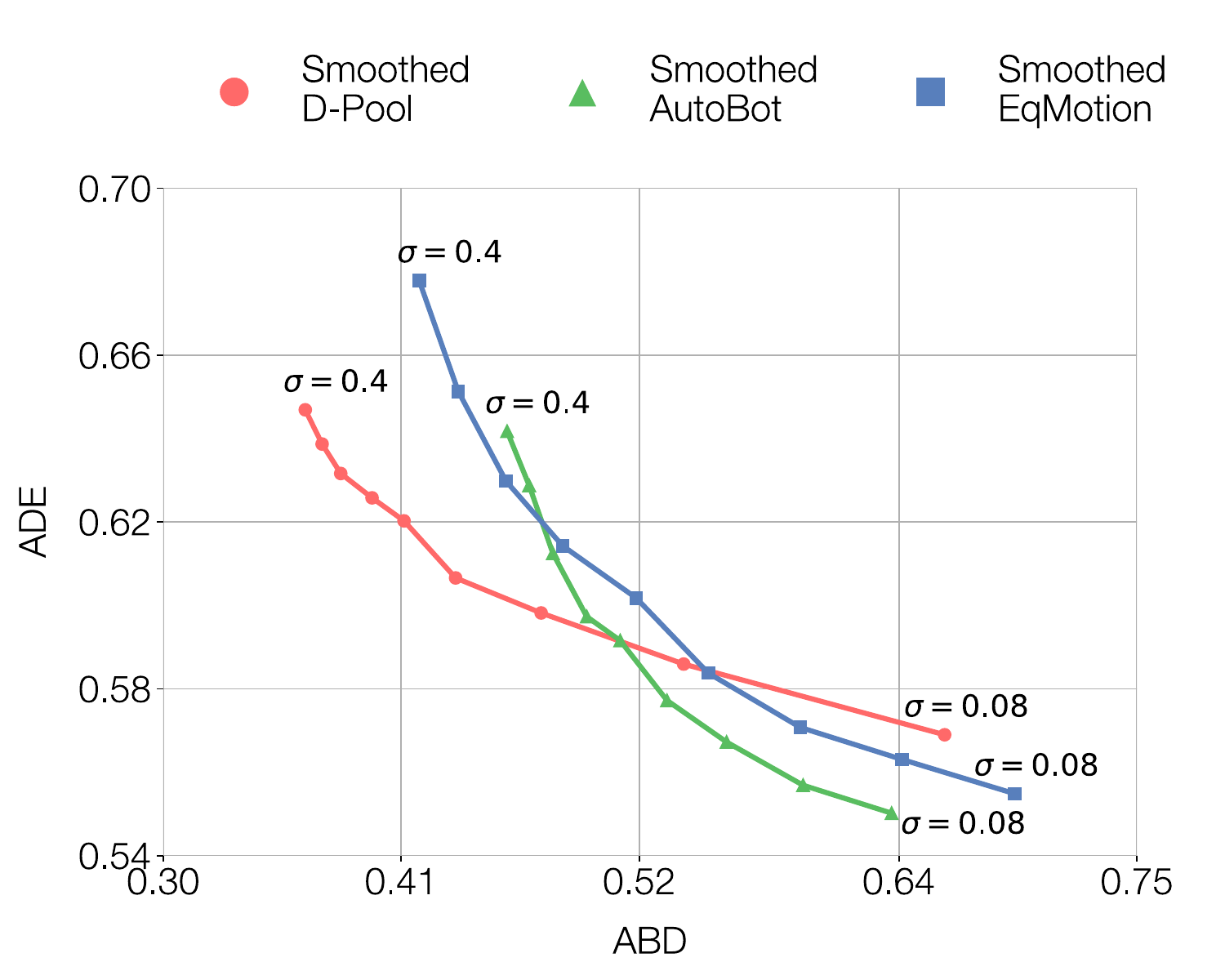} \includegraphics[width=0.99\columnwidth]{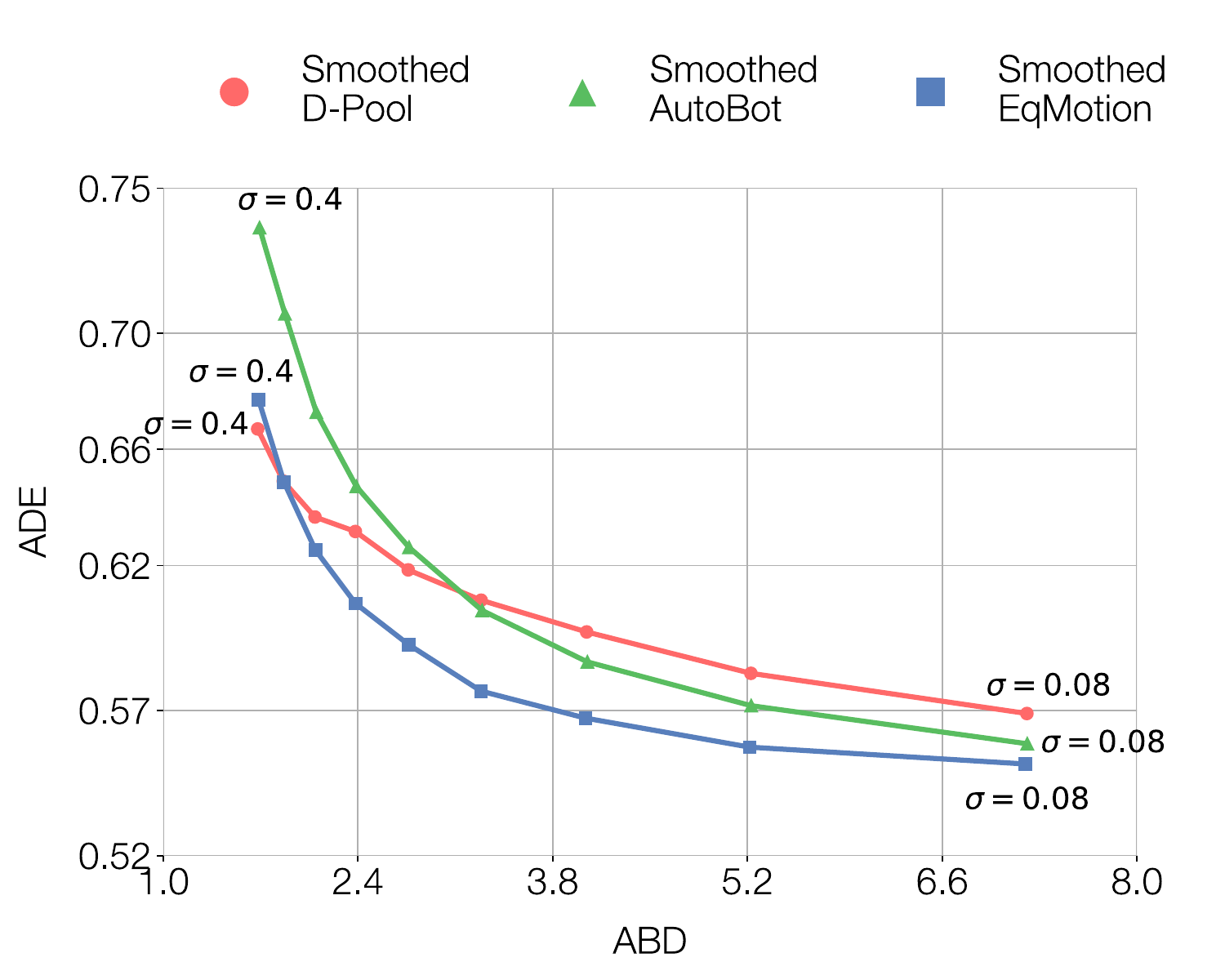}
    \caption{ADE against ABD for median and mean aggregations, respectively. The results are for different smoothed predictors and equally spaced $\sigma$ within $[0.08,0.4]$.
    The bottom left indicates the best performance. The conclusions are similar to the main paper.}
    \label{fig:mean_baselines_ade}
\end{figure*}

\begin{figure}[!t]
    \centering
    \includegraphics[width=0.99\columnwidth]{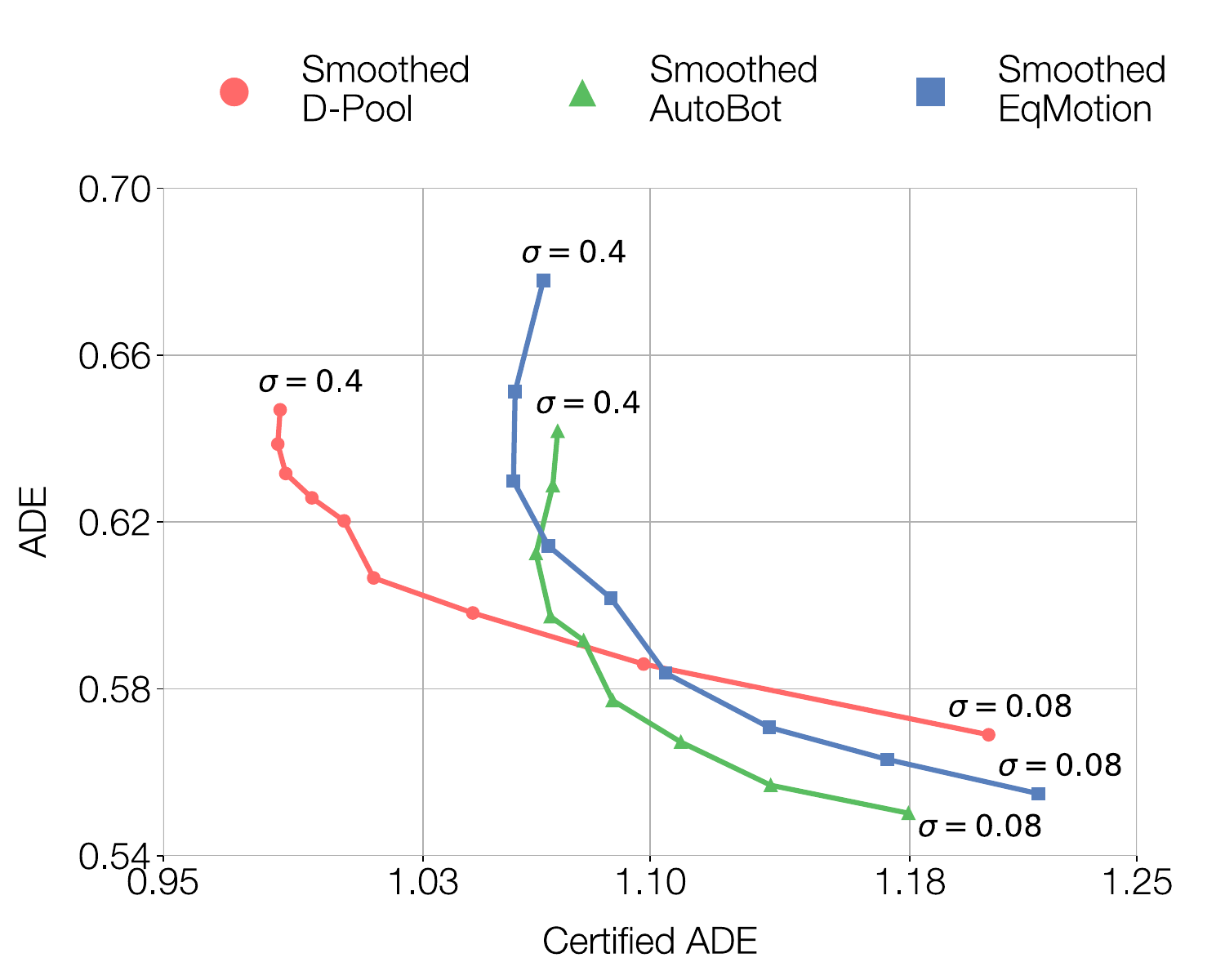}
    \caption{
    ADE against Certified-ADE.
    The results are for different smoothed predictors with median aggregation function and equally spaced $\sigma$ within $[0.08,0.4]$. The bottom left indicates the best performance. The conclusions are similar to the main paper.}
    \label{fig:certified_ade}
\end{figure}

In \Cref{sec:exp} of the paper, we mainly reported results in terms of FDE, FBD, and Certified-FDE due to space constraints. Here we provide the results in terms of ADE, ABD, and Certified-ADE in \Cref{fig:mean_baselines_ade} and \Cref{fig:certified_ade}.

\subsection{More qualitative results}
\label{sec:qualitative}

\begin{figure*}[!t]
\centering
    \centering
    \includegraphics[width=0.305\linewidth]{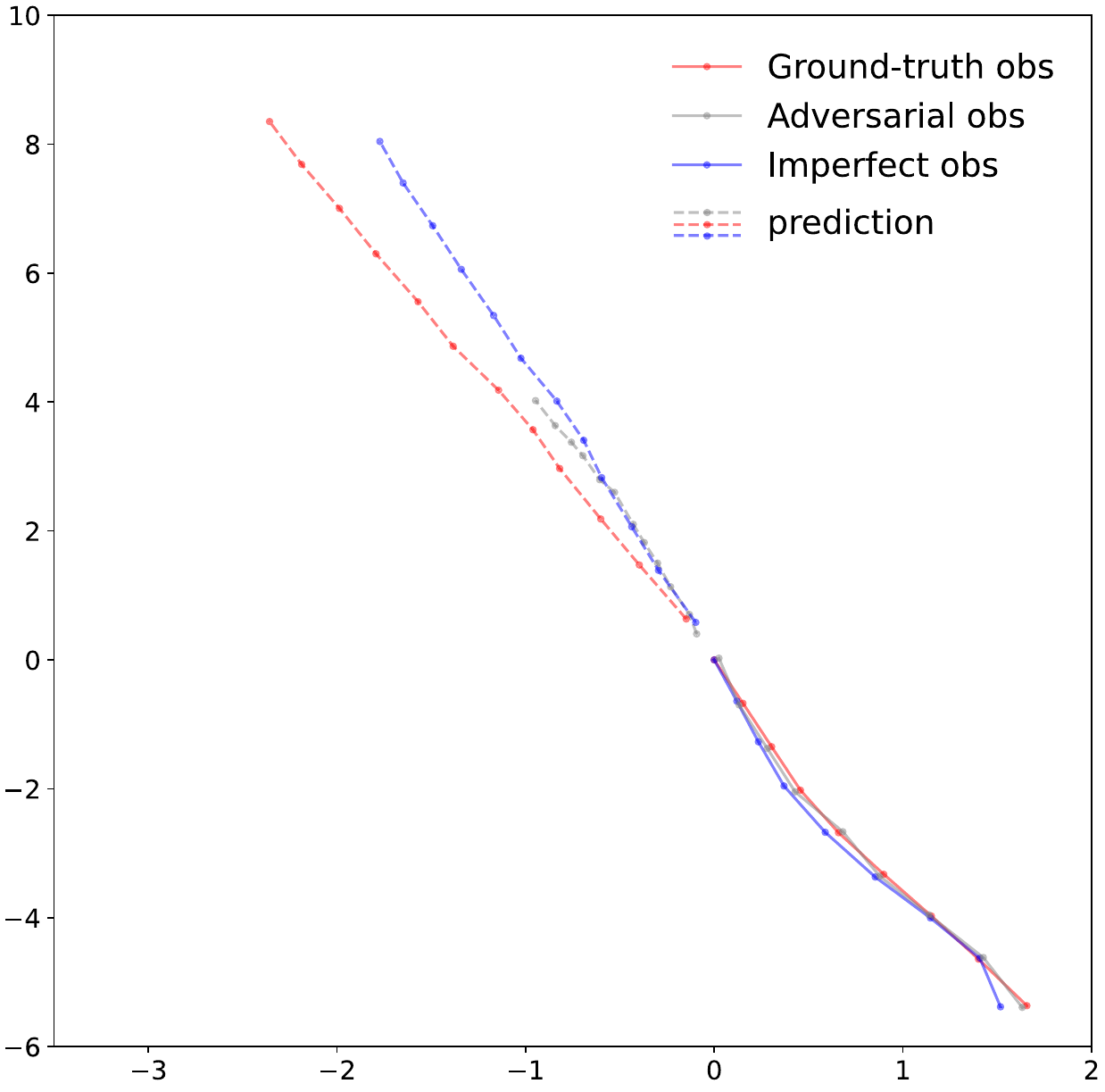}
    \includegraphics[width=0.31\linewidth]{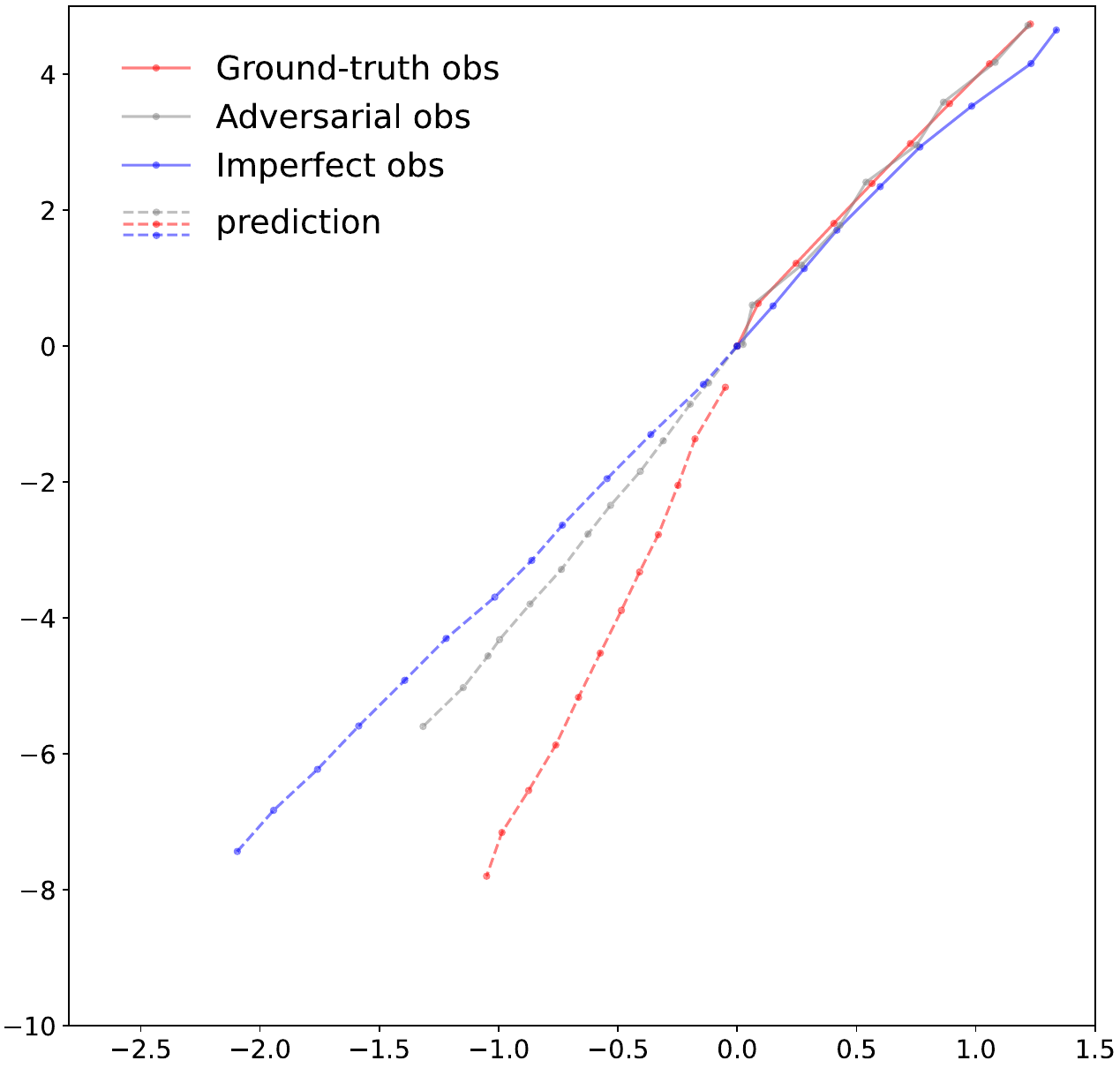}
    \includegraphics[width=0.3\linewidth]{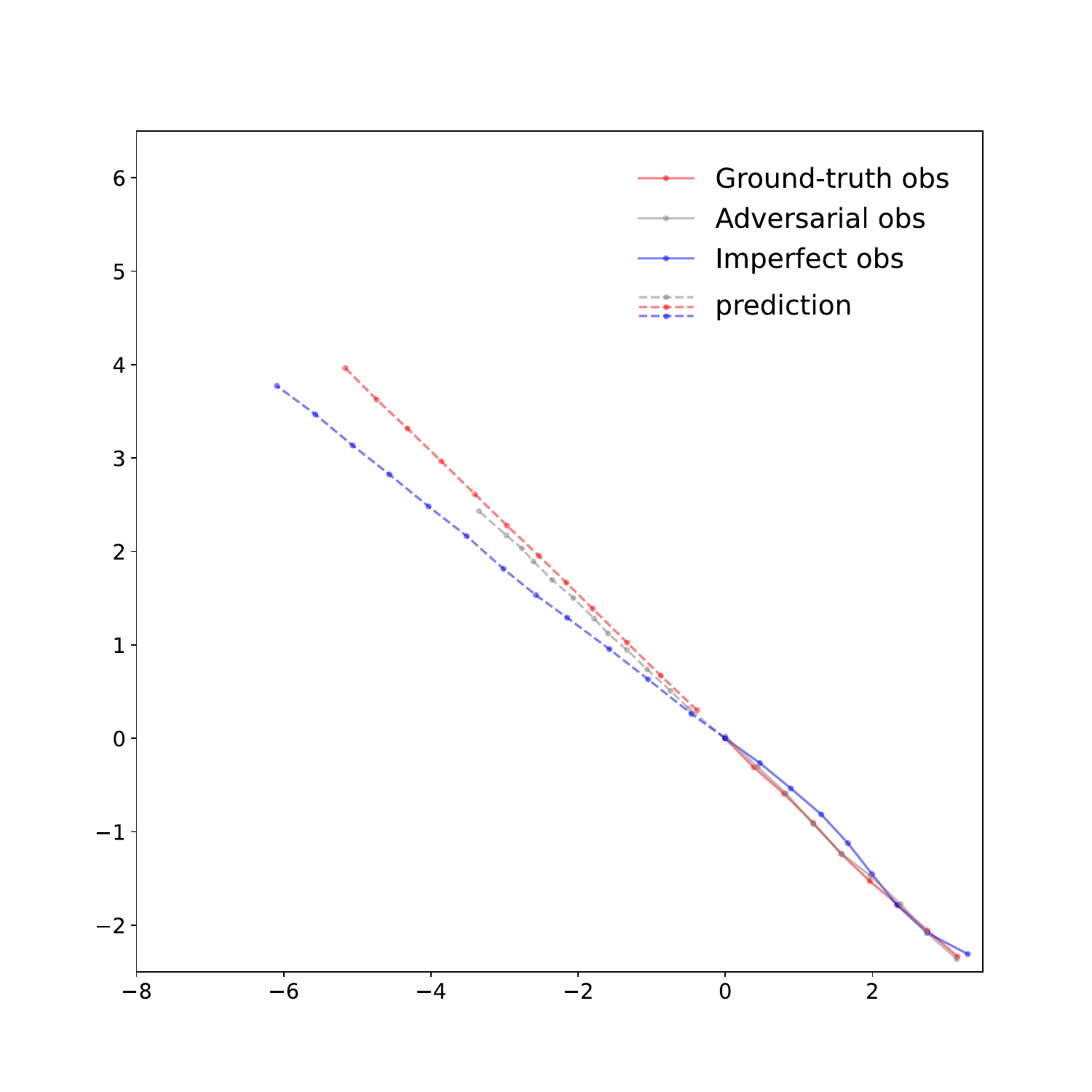}
    \\
    \includegraphics[width=0.305\linewidth]{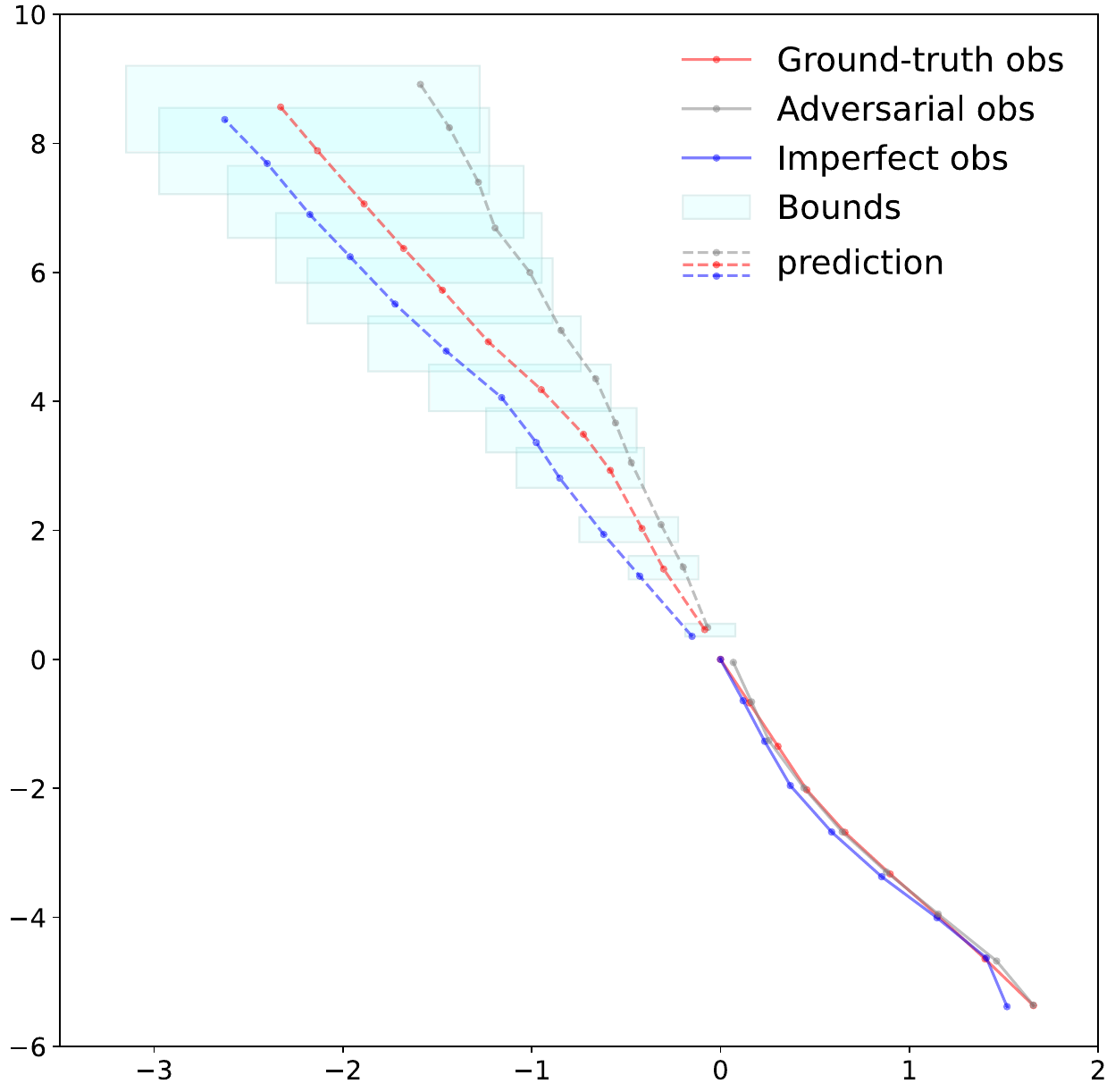}
    \includegraphics[width=0.31\linewidth]{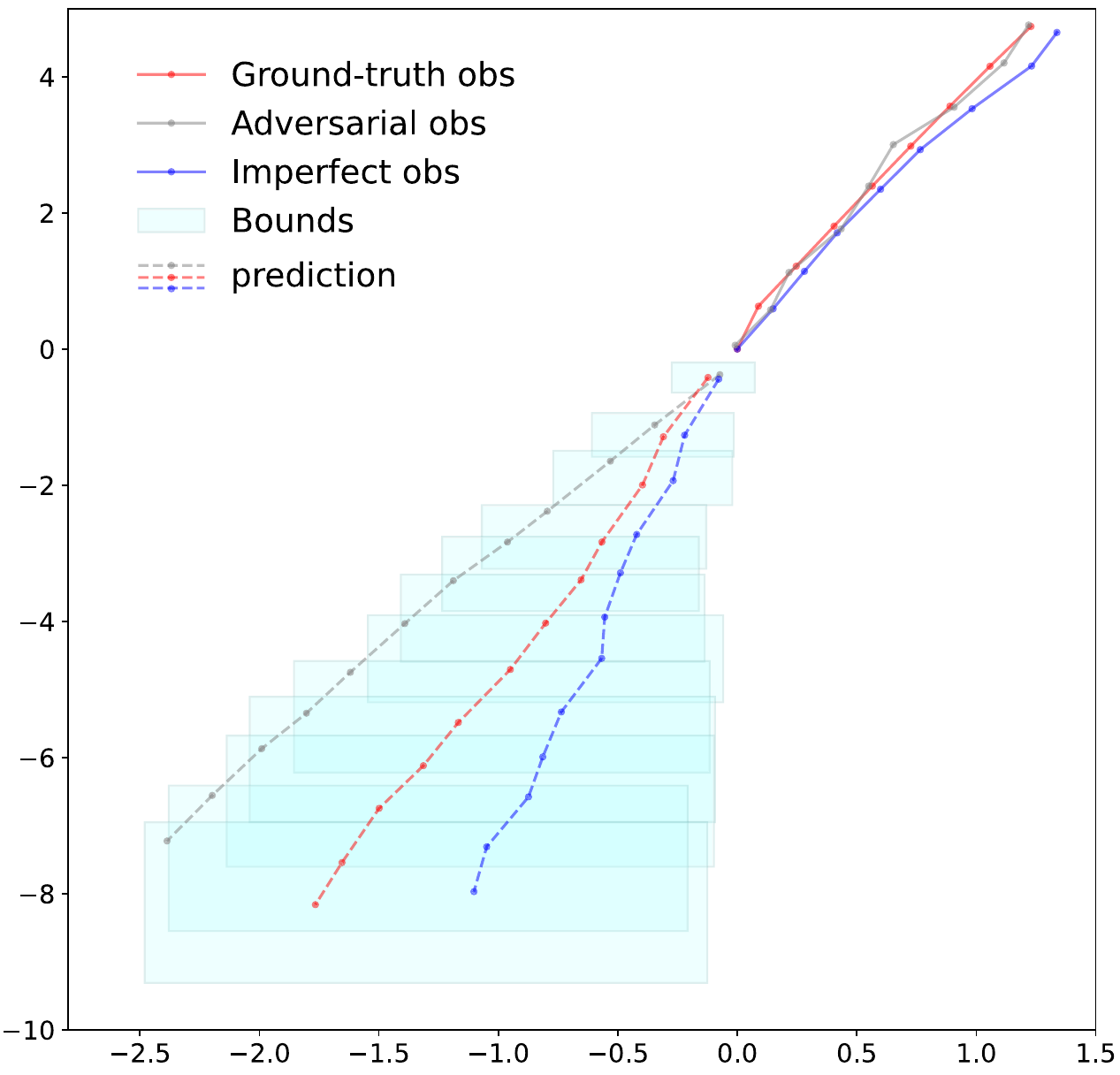}
    \includegraphics[width=0.3\linewidth]{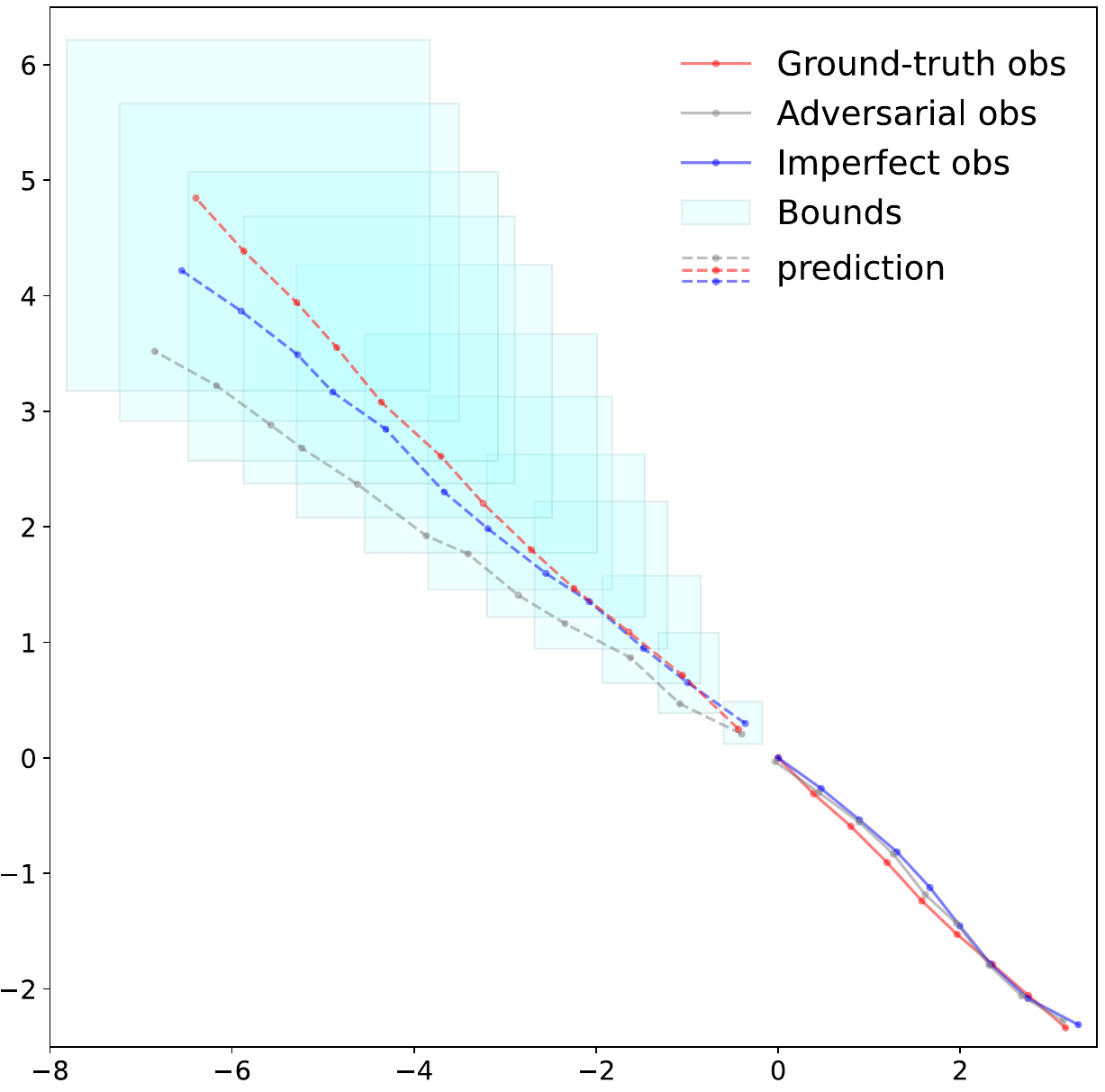} 
\caption{Comparing the performance of the original predictor (on the top) and the smoothed predictor (on the bottom). The red trajectories depict original observations, the blue trajectories represent predictions with imperfect observations coming from detection and tracking algorithms on real-world data, and the gray ones show the predictions given adversaries.}
\label{fig:attack_realword}
\end{figure*}

We showed a scenario in the main paper where we showcase the impact of an adversarial attack and imperfect observation on the performance of the predictor. Here, we provide more scenarios in \Cref{fig:attack_realword}. These results demonstrate that the models are vulnerable to different input noises, and certification can provide guaranteed robustness. 

\begin{figure*}[!h]
\centering
    \includegraphics[width=\linewidth]{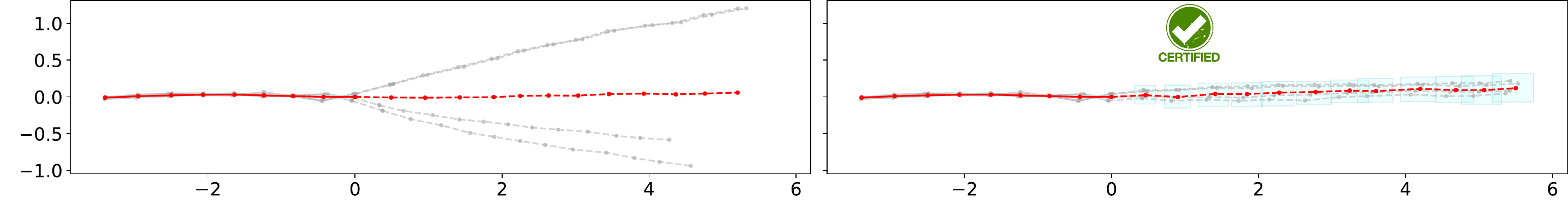}
\\
\centering
    \includegraphics[width=\linewidth]{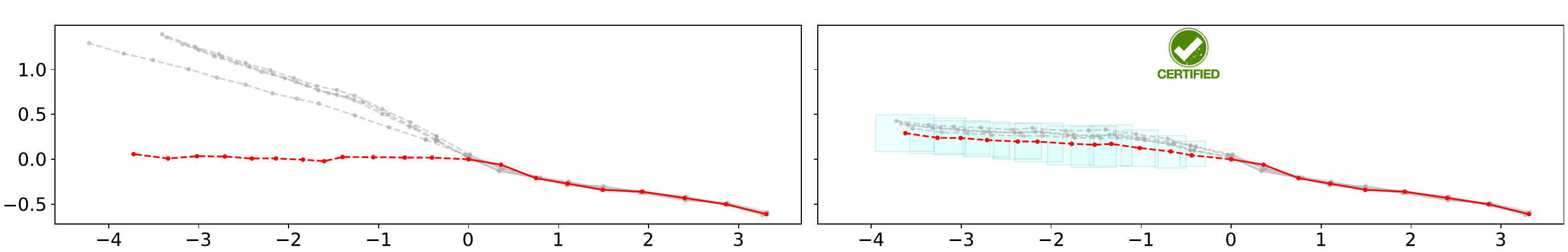}
\\
    \centering
    \includegraphics[width=\linewidth]{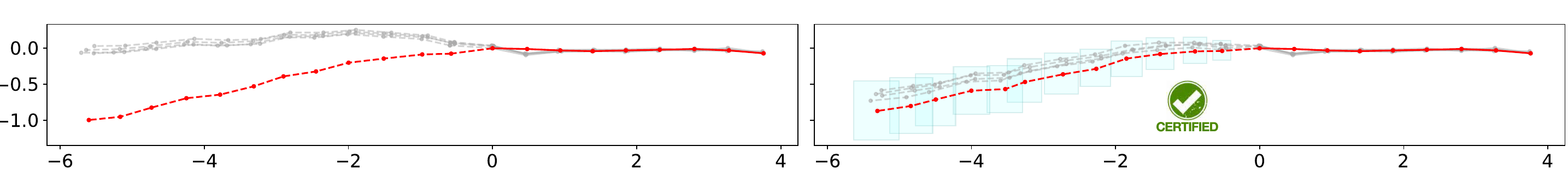}
\\
\centering
    \includegraphics[width=\linewidth]{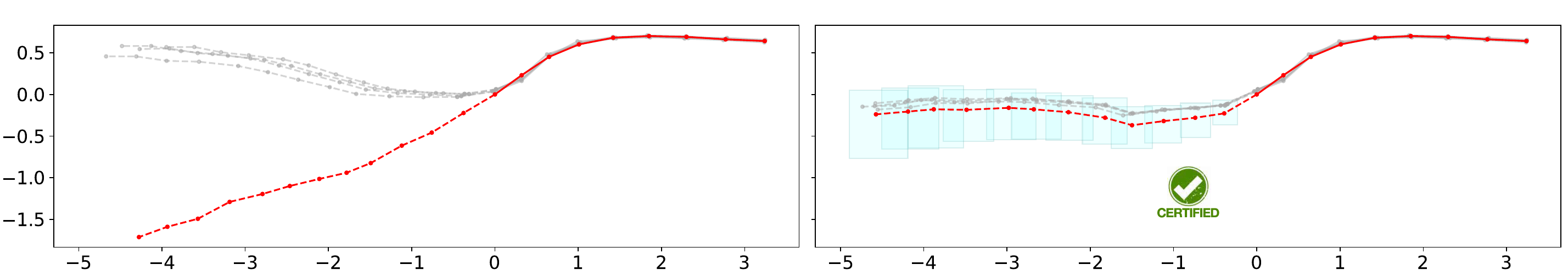}
    \caption{Qualitative results of the original predictor compared with the smoothed predictor. The red trajectories depict clean inputs and the corresponding predictions, and the gray trajectories represent noisy inputs and predictions. 
    The left part showcases the outputs of the original predictor, revealing unbounded predictions. In contrast, the right part demonstrates the outputs of the smoothed predictor, underscoring our ability to certify bounds on predicted outputs.}
\label{fig:supp_qual1}
\end{figure*}

In \Cref{fig:supp_qual1}, we show qualitative results of EqMotion and smoothed EqMotion. We generate multiple noisy inputs by adding random noise with a magnitude less than $0.1$ to an input trajectory and visualize the models' predictions. As evidenced, the original predictor yields highly variable outputs, however, the smoothed predictor predicts within the certified bounds.
It is important to note that the certified bounds are functions of the input; consequently, they are larger in some scenarios and smaller in others.

\begin{figure}[!t]
    \centering
    \includegraphics[width=\columnwidth]{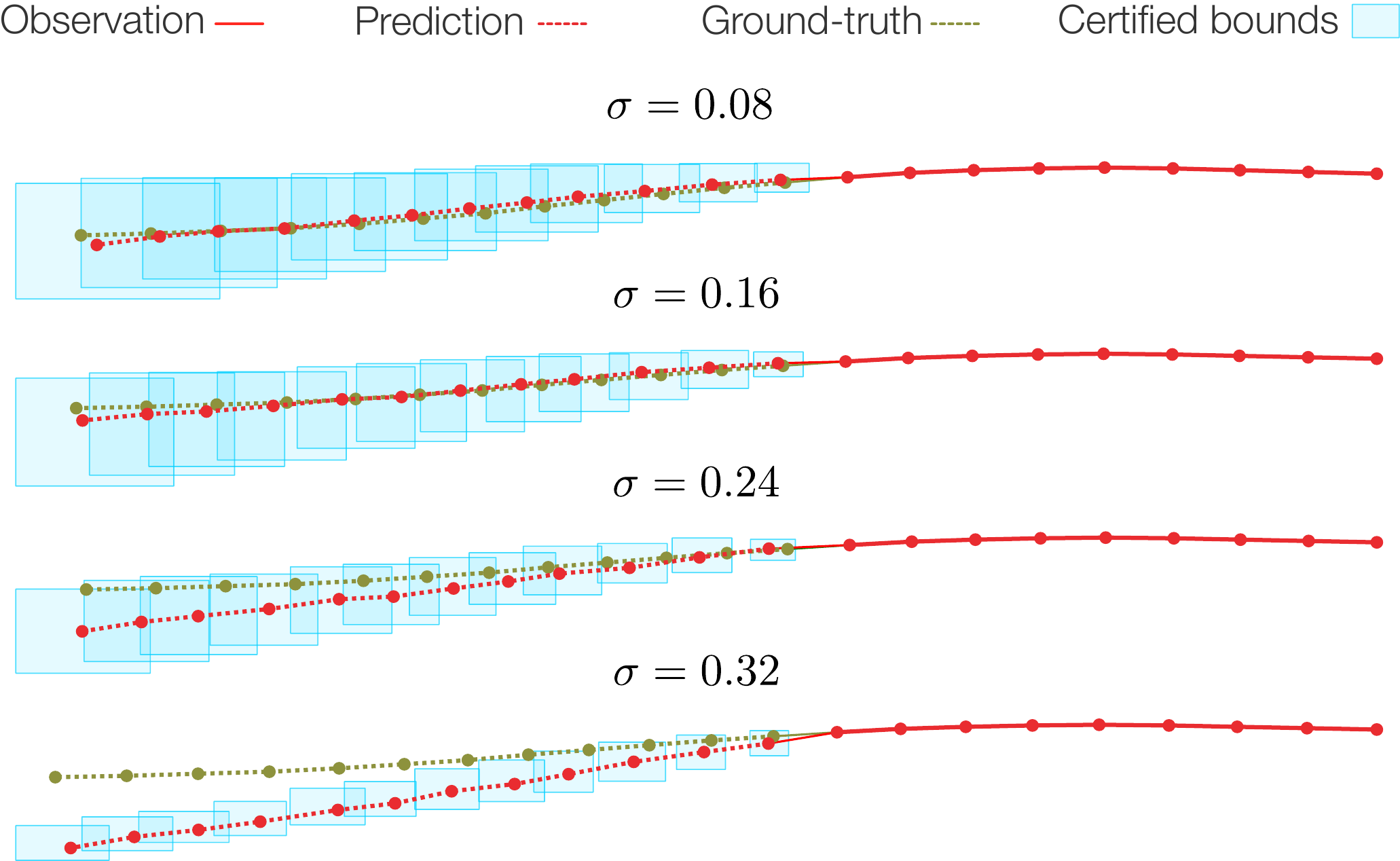}
    \caption{Qualitative results of our model for different values of $\sigma$. It shows the outputs of the smoothed EqMotion for one randomly selected data sample in the dataset.The ground-truth predictions are depicted in green, while the observation and the model's predictions are in red. 
    The figure shows that increasing $\sigma$ tightens the bound at the cost of dropping the accuracy.
    }
    \label{fig:qualitative_sigma}
\end{figure}

\Cref{fig:qualitative_sigma} presents qualitative results across varying $\sigma$ in the smoothed function. It demonstrates the trade-off between accuracy and the bound size. As the sigma value increases, the perturbation overwhelms the original input, resulting in a signal whose median aligns closely with the noise median, which is zero. Therefore, the bounds become tighter, but the accuracy drops. 
Among the various smoothing functions depicted, the one with $\sigma = 0.16$ appears to maintain a better balance, offering sufficiently tight bounds without significantly compromising accuracy, while the function with $\sigma = 0.32$ demonstrates relatively lower accuracy. 

\subsection{Certified bound per timestep}
\label{sec:timestep}

\begin{figure}[!t]
    \centering    \includegraphics[width=\columnwidth]{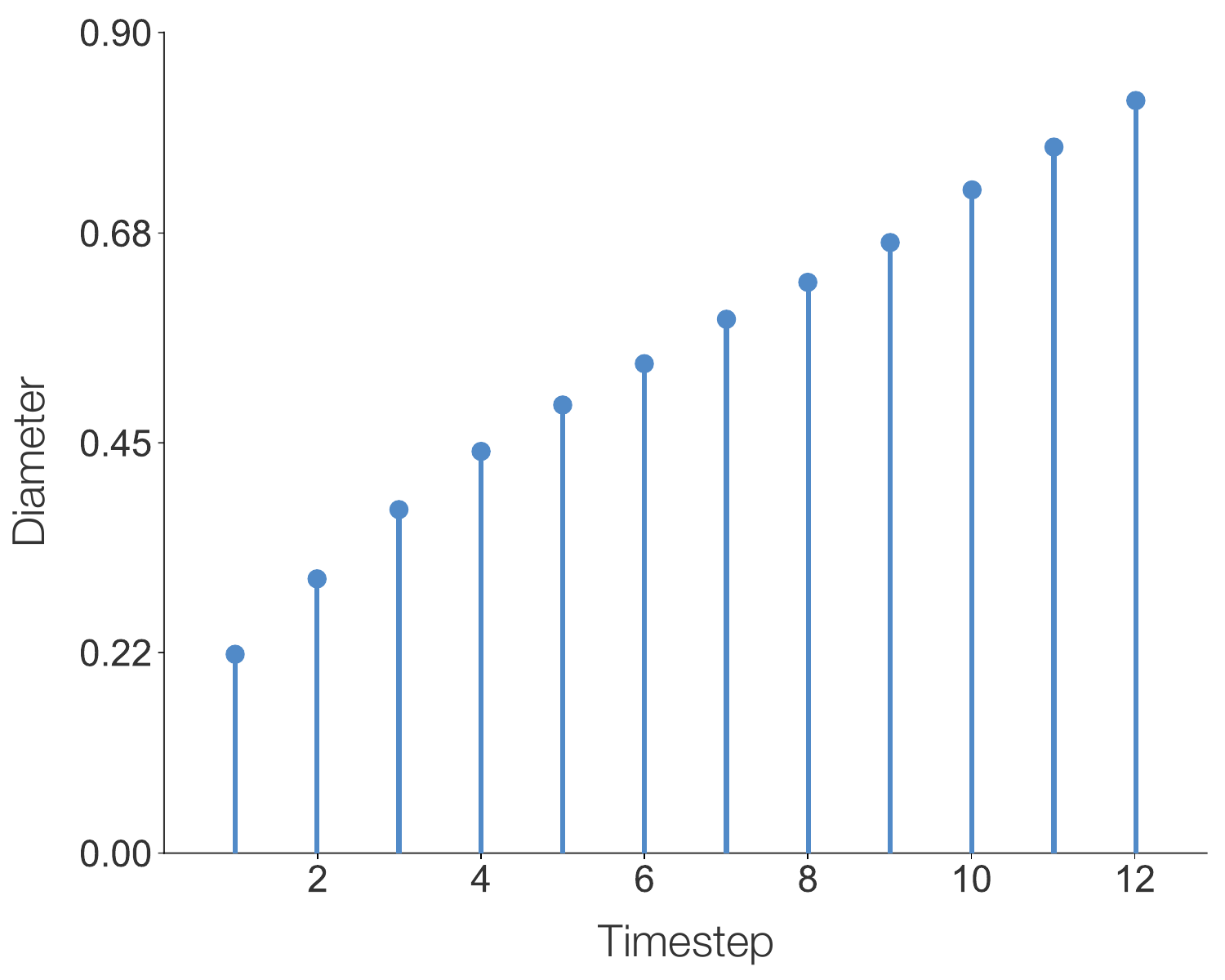}
    \caption{Certified bound per timestep. 
    We report the distance of the farthest point in the certified bound to the predicted trajectory for different timesteps as the bound diameter. Smoothed EqMotion with $\sigma=0.2$ is employed for this experiment.  It shows that later timesteps have higher bounds due to their larger output variation.}
    \label{fig:timestep}
\end{figure}

Until now, we have reported the final and average certified bounds. However, each prediction timestep has a different bound. \Cref{fig:timestep} shows the bounds for each timestep. As expected, later timesteps have larger bounds, correlating with their potentially greater variations.

\subsection{Analyzing the bounds with adversarial attacks}
\label{sec:adv_appendix}
Additionally, we conduct two experiments on a subset of Trajnet++ dataset to further investigate the certified bounds presented in the paper. 

We first compare the Certified-FDE of Smoothed EqMotion to those of the original EqMotion model. Remind that Certified-FDE is the guaranteed worst-case FDE happening given input deviations (we use worst-case FDE and Certified-FDE interchangeably in this subsection). However, there is no guarantee for the worst-case FDE of the original model. In order to determine a lower-bound for the worst-case FDE of the original model, we employ an adversarial attack. Note that it is a lower-bound since our attack is one potential attack and not necessarily the strongest possible attack, and the worst-case FDE could potentially be higher with other adversarial attack approaches.
We employ the PGD attack~\cite{madry2017pgdattack}, constrain the $L_2$ norm of perturbations to $0.1$, similar to the value of $R$ in our main experiments, and use a subset of trajnet++ dataset. The objective is to find perturbations that would increase the FDE for the EqMotion model. This attack demonstrated that applying adversarial perturbations could raise the FDE of EqMotion from $1.12$ to $1.73$.
On the other hand, the Certified-FDE for the Smoothed EqMotion 
is $1.87$. This shows that while a lower-bound for the worst-case FDE of the original model is $1.73$, the guaranteed worst-case FDE for the Smoothed predictor is $1.87$ which is within the same range but guaranteed. 
This means that any attack to the smoothed predictor, as long as it adheres to the $L_2$ norm constraint, will result in outputs that fall within the guaranteed worst-case FDE.

\end{document}